%% file: main.tex
\documentclass[11pt]{article}

\usepackage{iftex}
\ifPDFTeX
\fi

\usepackage[preprint]{acl}

\usepackage{times}
\usepackage{latexsym}

\ifPDFTeX
  \usepackage[T1]{fontenc}
\else
  \usepackage[UTF8,scheme=plain,fontset=fandol]{ctex}
\fi

\ifPDFTeX\usepackage[utf8]{inputenc}\fi
\usepackage{amsmath}
\usepackage{amssymb}

\usepackage{microtype}

\ifPDFTeX
  \usepackage{inconsolata}
\else
\fi

\usepackage{graphicx}
\usepackage{svg}
\usepackage[capitalize]{cleveref}
\usepackage{nicefrac}
\usepackage{booktabs}
\usepackage{multirow}
\usepackage{listings}
\usepackage{placeins}
\usepackage{stfloats}
\renewcommand{\floatpagefraction}{0.8}

\renewcommand{\dblfloatpagefraction}{0.8}

\lstdefinestyle{promptstyle}{
  basicstyle=\ttfamily\scriptsize,
  breaklines=true,
  breakatwhitespace=false,
  columns=fullflexible,
  frame=single,
  xleftmargin=0pt,
  xrightmargin=0pt,
  framesep=4pt,
  showstringspaces=false,
  keepspaces=true,
  breakindent=0pt,
  postbreak={}
}

\title{Diagnosing Task Insensitivity in Language Agents}
\author{
  Jingyu Liu\textsuperscript{1},
  Jiaen Lin\textsuperscript{3},
  Xiaopeng Wu\textsuperscript{2},
  Kehan Chen\textsuperscript{2},
  Chuan Yu\textsuperscript{2},
  Yong Liu\textsuperscript{1,4,5\ensuremath{\dagger}}\\
  \normalfont\small\textsuperscript{1}Gaoling School of Artificial Intelligence, Renmin University of China, Beijing, China\\
  \normalfont\small\textsuperscript{2}Taobao \& Tmall Group of Alibaba\\
  \normalfont\small\textsuperscript{3}School of Software, Tsinghua University, Beijing, China\\
  \normalfont\small\textsuperscript{4}Beijing Key Laboratory of Research on Large Models and Intelligent Governance\\
  \normalfont\small\textsuperscript{5}Engineering Research Center of Next-Generation Intelligent Search and Recommendation, MOE\\
  \normalfont\small\texttt{liujy1016@ruc.edu.cn}
}

\input{tables/main_response_rows}
\input{tables/group_rates}
\input{tables/replacement_endpoints}
\input{tables/attention_granularity_rows}
\input{tables/selective_rollout_tables}

\input{tables/task_checking_tables}

\begin{document}
\maketitle

\begin{abstract}

Language agents select actions based on task descriptions,
environmental observations, and interaction histories.
We investigate whether task requirements reliably constrain these
decisions when the surrounding context already suggests how to proceed.
Across five models and three interactive environments, we
find that models, including GPT-5.6-Sol, can continue to act
when task descriptions are replaced with the meaningless
placeholder \texttt{xxxx}, even when prompted to ask for clarification.
Action generation is more frequent in states that suggest a concrete next operation,
although these cues do not establish whether that operation serves the user's goal.
These findings suggest a possible training-based explanation: contextual cues can
make demonstrated responses predictable, allowing models to fit them with limited
reliance on task requirements. Tracking seven models during fine-tuning,
we observe an overall decrease in attention to task-description tokens and
an increase in attention to current observations. To strengthen task dependence,
we introduce Task-Perturbed Negative Log-Likelihood (TPNLL), which
contrasts the prediction costs of the same response under different
task descriptions while holding the non-task context fixed.
TPNLL increases attention to task-description tokens and improves
complete-episode success in out-of-distribution settings.
\end{abstract}

\section{Introduction}

\begin{figure}[!t]
  \centering
  \includegraphics[width=\columnwidth,trim=0 0 0 6bp,clip]{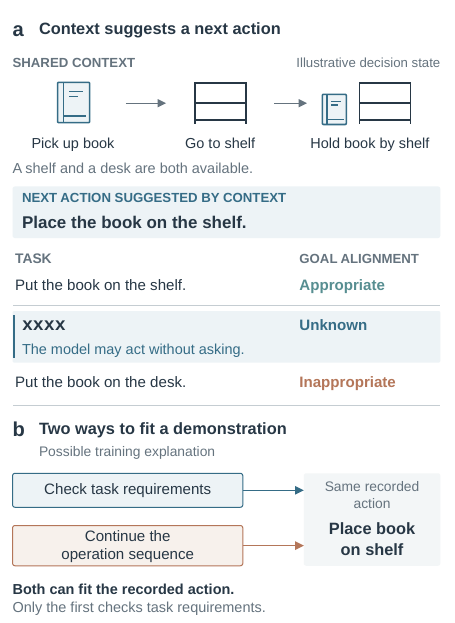}
  \caption{Task dependence in action selection and training.
  (a) Context suggests placing a book on the shelf, but whether this action
  serves the goal depends on the task.
  (b) Two possible routes to predicting the same demonstrated action:
  checking task requirements or continuing the operation sequence suggested
  by context.}
  \label{fig:context_task_overview}
\end{figure}

Language agents support a wide range of tasks, including software
development \citep{yang2024sweagent} and computer use
\citep{xie2024osworld}. Their reliability depends on whether
their actions satisfy the user's task requirements
(Figure~\ref{fig:context_task_overview}).
Public reports of agents acting beyond explicit task boundaries
illustrate the practical importance of maintaining task constraints
throughout execution
\citep{garcia2026geminireview,hyphymike2026codexdeployment}.
We find that current models sometimes continue operations suggested by
observations and interaction history without adequately using the task
description. We study \emph{task dependence}: how reliably task requirements
constrain action selection when the surrounding context already suggests
how to continue.

To test whether continued action requires interpretable goal information,
we evaluate five existing models, including
GPT-5.6-Sol, in
ALFWorld \citep{alfworld}, ScienceWorld \citep{scienceworld}, and WebShop
\citep{webshop}. We replace the task description with the meaningless placeholder
\texttt{xxxx}, keeping the remaining context unchanged, requesting analysis
before the final response, and explicitly allowing clarification questions.
All five models still generate
environment actions in some states without first obtaining an
interpretable goal. These results show that a usable task description is
not always a prerequisite for continued operation.

We further find that action generation is generally more frequent when
context suggests a concrete next operation. In ALFWorld, for example,
picking up an object and arriving at a receptacle can suggest opening
the receptacle or placing the object. Across ALFWorld states with concrete
follow-up cues, GPT-5.6-Sol produces environment actions in 99.7\% of
responses, versus 28.9\% in states with an unresolved operational choice.
Operational context can make a next action predictable without
establishing whether that action serves the user's goal.

This pattern suggests a possible training explanation. The model might
learn to rely on the context to make decisions.
To investigate how this dependence develops during training,
we track the fine-tuning process of seven publicly available models.
We find that as training progresses, attention to task-description tokens
in the prompt gradually declines, while attention to the current
observation increases.
This directly shows that training reduces reliance on task descriptions.

To make predictions depend on task requirements within the same context,
we introduce
\textbf{Task-Perturbed Negative Log-Likelihood (TPNLL)}. It contrasts
the same recorded response under the original task and a similar but
distinct replacement, keeping the non-task context fixed. An
auxiliary penalty encourages predictions to remain responsive to this
task change, using the model at initialization as a reference.
TPNLL improves complete-episode out-of-distribution (OOD) success while making
the model more sensitive to the task description.

Our contributions are as follows:
\begin{itemize}
  \item We show that current models can make concrete action choices
  based on context, without an interpretable task, and relate this
  continuation to contextual cues that suggest a next operation.
  \item We show that task dependence can weaken during training with
  complete task descriptions, with attention to task-description tokens
  gradually decreasing as training progresses.
  \item We introduce initialization-calibrated TPNLL to strengthen
  task dependence through contrasts between response likelihoods under
  different task conditions, and demonstrate improvements in OOD task success.
\end{itemize}

\section{Related Work}

\paragraph{Task dependence in language agents.}
Clarification and abstention studies reveal a gap between recognizing a
problem and changing behavior: models can detect ambiguity without asking
questions \citep{su2026knowing,matsnev2026uncertainty} or acknowledge constraints yet continue
executing \citep{liu2026agentabstain}. 
Instruction-corruption and contrast-set evaluations likewise probe
whether behavior follows task requirements
\citep{parekh2024instruction,anwar2024contrast,hou2026langgap}.
We examine which contextual cues sustain action when task information is
unavailable, then investigate how dependence on task descriptions changes
during training.

\paragraph{Training and shortcut learning.}
Observational and temporal correlations can provide predictive shortcuts
in supervised and imitation learning
\citep{geirhos2020shortcut,dehaan2019causal,wen2020copycat}.
Instruction-tuning studies report comparable performance after removing
task-definition semantics while retaining output-space information
\citep{kung2023instruction}, and LangForce find VLMs could neglect the instruction because
the observations already predict the instruction \citep{lian2026langforce}. And \citet{liu2026cotgains}
shows that the model might learn to predict the answer without thinking process.
Building on these accounts, we track task dependence across language-agent
training checkpoints. We observe declining attention to task descriptions
alongside more frequent repetition of unsuitable actions after task
replacement, and introduce TPNLL to strengthen task dependence.

\section{Action Selection with Insufficient Task Information}
\label{sec:measurement}
\label{sec:placeholder_behavior}
\label{sec:current_agent_setup}
\label{sec:fixed_context}

Task descriptions specify which actions serve the user's goal when
several actions are plausible in the same environment. Clarification
studies examine whether models recognize ambiguity and request missing
information \citep{zhang2024clamber,zhang2024askbeforeplan}. Our focus is
how task information constrains action selection: whether agents continue
acting without it and how contextual cues shape that continuation.

\begin{figure*}[!t]
  \centering
  \includegraphics[width=\textwidth]{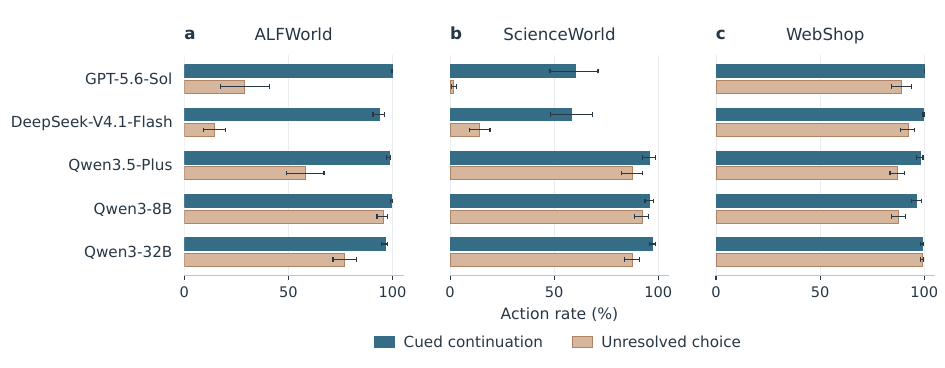}
  \caption{Behavior under TASK = \texttt{xxxx} in cued-continuation and
  unresolved-choice states. Paired bars compare the two groups for each
  model within each environment. Bars show ordinary-action rates among all responses;
  error bars give pointwise 95\% state-bootstrap intervals.}
  \label{fig:context_next_step_groups}
\end{figure*}

\subsection{Acting without an Interpretable Task}
\label{sec:acting_without_task}

We sample decision states from real interaction trajectories, masking
explicit task descriptions and checking for residual goal disclosure
(Appendix~\ref{app:existing_state_selection}).

We then replace the task description with \texttt{xxxx}, leaving the history,
current observation, available actions, original operating instructions,
and step budget unchanged. The uninterpretable placeholder makes the
absence of usable task information explicit. We ask models to provide
analysis in \texttt{<think>} before returning an environment action in
\texttt{<action>} or a clarification question in \texttt{<inquiry>}.
We use 150 states each
in ALFWorld and ScienceWorld and 100 in WebShop, with ten responses per
state and model. We evaluate GPT-5.6-Sol, DeepSeek-V4.1-Flash, Qwen3.5-Plus,
and Qwen3-8B/32B \citep{yang2025qwen3}; sampling details appear in
Appendix~\ref{app:existing_agent_evaluation}.
The interaction history retains previous actions and environment feedback
but excludes earlier reasoning, preventing task information from being
disclosed through prior reasoning text.

We classify responses as Inquiry (requests for task information), Act
(other parsed environment actions), or Other (unclassified or empty
outputs), parsing the final response separately from the analysis.

\begin{table}[!htbp]
  \centering\small
  \setlength{\tabcolsep}{5pt}
  \begin{tabular}{@{}lccc@{}}
    \toprule
    Model & Inquiry & Act & Other\\
    \midrule
    \ContextHistoryFullRows
    \bottomrule
  \end{tabular}
  \caption{Response proportions (\%) under TASK = \texttt{xxxx}, with
  native history and current observation retained.}
  \label{tab:placeholder_behavior}
  \label{tab:current_agent_responses}
\end{table}

Table~\ref{tab:placeholder_behavior} shows that an uninterpretable TASK
does not necessarily prevent action generation. GPT-5.6-Sol, for example,
produces ordinary environment actions in 81.3\%, 36.7\%, and 94.6\% of
responses in ALFWorld, ScienceWorld, and WebShop, respectively; the other
models also act directly in some states. These results show that a usable
task description is not always a prerequisite for continued operation.
The analysis in Appendix~\ref{app:task_checking} further shows that some
responses explicitly acknowledge the missing task information and still
proceed, while others act without expressing such acknowledgment.

\subsection{Context-Guided Action Continuation}
\label{sec:context_continuation}

Context can suggest a concrete next action without establishing whether
it serves the user's task. In WebShop, completed attribute selections
suggest purchase without confirming that the selected product meets the
user's requirements. We test whether such cues are associated with more
frequent action generation when the task description is unavailable.

We compare two groups of states from Table~\ref{tab:placeholder_behavior},
defined by their visible history, current observation, and action menu:
\emph{cued continuation} and \emph{unresolved choice}. In
\emph{cued continuation}, these records support one or more concrete
follow-up operations. In \emph{unresolved choice}, an object, destination,
attribute, or procedure remains to be selected. For example, WebShop
states with all visible attributes selected suggest purchase, whereas
states with an unselected attribute leave a choice unresolved. We assign
groups using environment-specific rules and the review procedure in
Appendix~\ref{app:context_continuation}. Both groups retain native history
and observation with TASK set to \texttt{xxxx}; the distinction concerns
the next operation, not whether the task is known.

Figure~\ref{fig:context_next_step_groups} shows generally higher action
rates in cued-continuation states. Thus, agents are more likely to proceed
without first obtaining usable task requirements when the visible context
supports concrete follow-up operations.

\section{Task Sensitivity during Training}
\label{sec:training}

The context-guided continuation observed above motivates a hypothesis
about training: models may learn to predict the next action from the
surrounding context with limited reliance on the task description,
reducing their sensitivity to task requirements.

\subsection{Attention Allocation during Training}
\label{sec:attention}

\begin{figure*}[!t]
  \centering
  \includegraphics[width=\textwidth,trim=0 0 0 5bp,clip]{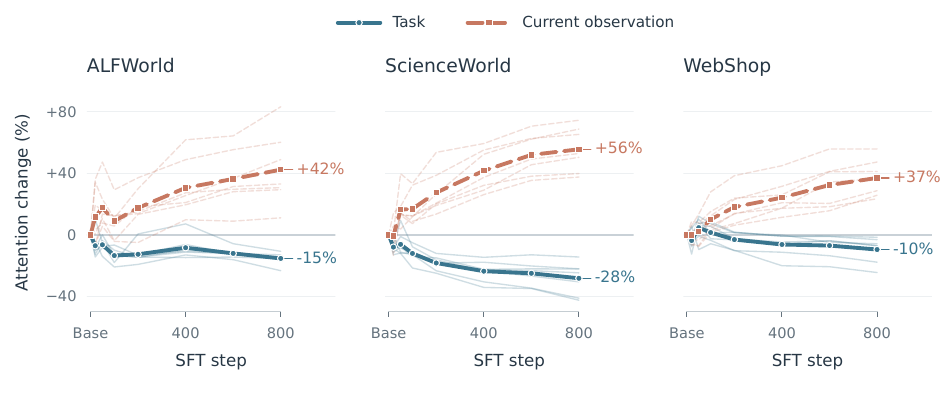}
  \caption{Relative change (\%) in per-token attention density from
  initialization (Base). Light curves: individual models; bold: their
  equal-weight mean.}
  \label{fig:response_attention_vanilla}
\end{figure*}

To study attention to task descriptions, we track Transformer attention
\citep{vaswani2017attention} during SFT. We fine-tune seven models on
interaction trajectories with complete task descriptions for 800 optimizer
updates, using a global batch size of 8.
For each environment, we measure attention on 500 fixed prompt--response
pairs containing complete original tasks.
We additionally train Qwen3-4B with reinforcement learning (RL) to test
whether a different training objective produces similar changes
(Appendix~\ref{app:rl_evaluation}).

We evaluate Qwen3-4B/8B/14B/32B \citep{yang2025qwen3},
Llama-3.1-8B-Instruct \citep{grattafiori2024llama3},
Llama-3.2-3B-Instruct \citep{meta2024llama32}, and
Ministral-8B-Instruct-2410 \citep{mistral2024ministraux} before SFT (Base)
and at steps 20, 50, 100, 200, 400, 600, and 800 of environment-specific SFT.

Formally, let $Q_i$ be the response-token queries for case $i$, let $K_i$
be all its task-description key tokens, let $\mathcal L$ and $\mathcal H$
denote the sets of layers and attention heads, and let
$\alpha^{(i)}_{\ell,h}(q,k)$ be the attention weight at layer $\ell$ and
head $h$. We compute
\begin{equation}
  T_i =
  \frac{1}{|Q_i|\,|\mathcal L|\,|\mathcal H|\,|K_i|}
  \sum_{\substack{q\in Q_i,\;\ell\in\mathcal L\\
                  h\in\mathcal H,\;k\in K_i}}
  \alpha^{(i)}_{\ell,h}(q,k).
  \label{eq:main_task_attention_density}
\end{equation}
Thus, $T_i$ is the mean attention received by a task token from
response-token queries. We refer to this per-token attention density as
\emph{task attention}. Teacher-forcing the same prompt--response pairs
at every checkpoint controls for changes in response content, length,
and case composition.

\begin{figure*}[!t]
  \centering
  \includegraphics[width=\textwidth,trim=0 5bp 0 8bp,clip]{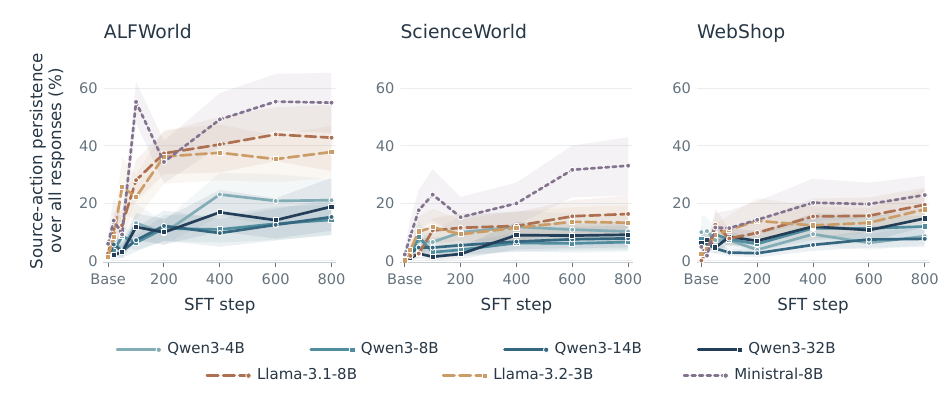}
  \caption{Source-action persistence over all sampled responses under task replacement during SFT.
  Base denotes initialization before SFT; subsequent points are SFT checkpoints.
  Shaded bands show pointwise 95\% source-task cluster-bootstrap intervals.}
  \label{fig:replacement}
\end{figure*}

Relative to Base, Figure~\ref{fig:response_attention_vanilla} shows an
overall decline in task-token attention and an increase in
current-observation attention during SFT. Complete task information remains
available in every input, yet receives less direct attention as training
progresses. These changes suggest a training route to context-guided
continuation: information that makes the next operation predictable
receives increasing attention, while the explicit goal receives less.

The decline extends across model components: a majority of layer--head
units show lower task attention at step 800 than at initialization in
20 of the 21 model--environment groups
(Appendix~\ref{app:attention_granularity}).
Vanilla RL on Qwen3-4B shows the same endpoint pattern in all three
environments: lower task-token attention and higher current-observation
attention (Appendix~\ref{app:rl_attention_success}).

\subsection{Action Persistence after Task Replacement}
\label{sec:explicit_replacement}

Tokens following the task description can themselves encode task
information, allowing the model to use it indirectly even when direct
attention to task tokens decreases. We therefore test \emph{task
sensitivity}: whether action selection responds to changes in task
requirements, and how this responsiveness changes during training.

We examine this question through task replacement:
we hold the non-task context fixed and substitute a task for which the
recorded action is unsuitable. A task-sensitive model should adjust its
action to the new requirement.

For each recorded state, we construct a complete replacement task from
the evaluation task families. Native environment replay
validates a feasible continuation for the new
goal. The non-task prompt remains fixed. Goal-specific checks and human review verify
that the source action selects an incompatible object, destination, or
option, or bypasses required processing.
Appendix~\ref{app:task_replacement} gives the full construction and
validation criteria.

Let $a_{\mathrm{src}}$ be the next action recorded in the original
trajectory at the evaluated state, fixed across all checkpoints. We sample five
responses per evaluation case under the replacement task and measure
\emph{source-action persistence}, using all sampled responses as the denominator:
\begin{equation}
  A_{\mathrm{all}}
  = \frac{\#\text{source-action matches}}
         {\#\text{sampled responses}}.
  \label{eq:all_persistence}
\end{equation}

A higher value means that the model more frequently
retains the recorded action that is unsuitable for the new task.
For illustration, suppose the recorded action selects blue on a product
page. If the task now requests red while the page remains unchanged,
selecting blue again counts as source-action persistence.

As shown in Figure~\ref{fig:replacement}, source-action persistence
gradually increases as training progresses.
In most settings, later checkpoints therefore more frequently
reproduce the recorded action despite its incompatibility with the new
requirement.
This shows that the model indeed becomes less sensitive to task descriptions.
Qwen3-4B also shows higher final persistence after Vanilla RL
(Appendix~\ref{app:rl_replacement}).

To examine whether this pattern persists with more diverse training tasks,
we train Qwen3-8B on a mixture of all three environments. At later
checkpoints, source-action persistence is higher and task attention lower
than at earlier checkpoints
(Appendix~\ref{app:mixed_environment_sft}). The pattern
thus persists when training spans multiple environments.

\begin{table}[!tb]
  \centering\small
  \setlength{\tabcolsep}{4pt}
  \renewcommand{\arraystretch}{1.12}
  \begin{tabular}{@{}p{\dimexpr\columnwidth-8pt\relax}@{}}
    \toprule
    \textbf{Replacement task}\\
    \texttt{put two cellphone in safe.}\\
    \textbf{State (summary)}\\
    Holding cellphone 1, at drawer 1; navigation to safe 1 is available.\\
    \midrule
    \textbf{Part of the model response}\\
    \texttt{<thinking>Task: put two cellphone in safe.} \ldots
    \texttt{</thinking>}\\
    \texttt{<action>move cellphone 1 to drawer 1</action>}\\
    \midrule
    \textbf{Verified continuation (first three actions)}\\
    \texttt{go to safe 1}\\
    \texttt{open safe 1}\\
    \texttt{move cellphone 1 to safe 1}\\
    \bottomrule
  \end{tabular}
  \caption{Correct task restatement with source-action persistence in
  ALFWorld after task-restatement SFT (Qwen3-8B, step 800). Only part of
  the response is shown; the state is summarized. Appendix~\ref{app:restatement_case}
  describes the case and its native replay verification.}
  \label{tab:task_restatement_example}
\end{table}

\paragraph{Explicit task restatement.}
Does the problem persist when training demonstrations explicitly restate
the task before choosing an action? We collect GPT-5.4 trajectories whose
responses begin by restating the task inside the reasoning block.
We train Qwen3-8B from its base initialization for 800 steps in
each environment and evaluate the same complete task-replacement pairs,
with five samples per input.

We find that attention to task descriptions still decreases and
source-action persistence remains. Table~\ref{tab:task_restatement_example}
illustrates this pattern: the model
restates that two cellphones belong in the safe, then places the held
cellphone in the drawer, retaining the source action.
Appendix~\ref{app:task_restatement}
reports the full protocol, curves, and attention measurements.

Together, these evaluations distinguish having task information from
using it to constrain actions. Agents may act without an interpretable
task or retain an unsuitable action after receiving a complete new task.

\begin{table*}[!tbp]
  \centering
  \small
  \setlength{\tabcolsep}{4pt}
  \begin{tabular}{@{}lcccccc@{}}
    \toprule
    & \multicolumn{4}{c}{Qwen3-8B} & Qwen3-14B & Llama-3.1-8B\\
    \cmidrule(lr){2-5}\cmidrule(lr){6-6}\cmidrule(lr){7-7}
    Method & ALF & Sci & Web & Avg. & Avg. & Avg.\\
    \midrule
    Vanilla SFT & $54.9{\pm}4.4$ & $25.4{\pm}0.7$ & $47.3{\pm}2.8$
      & 42.5 & 47.1 & 36.8\\
    NLL Difference & $55.5{\pm}3.2$ & $29.3{\pm}6.4$ & $46.6{\pm}2.2$
      & 43.8 & 44.9 & 37.2\\
    Margin Hinge & $58.6{\pm}8.6$ & $30.9{\pm}3.9$ & $47.0{\pm}3.9$
      & 45.5 & 43.9 & 39.0\\
    Unlikelihood & $55.5{\pm}7.0$ & $26.2{\pm}5.1$ & $48.8{\pm}2.3$
      & 43.5 & 44.8 & 34.8\\
    \textbf{TPNLL} & $\mathbf{59.4{\pm}2.8}$ &
      $\mathbf{32.3{\pm}0.5}$ & $\mathbf{51.0{\pm}2.5}$
      & \textbf{47.6} & \textbf{52.6} & \textbf{40.1}\\
    \bottomrule
  \end{tabular}
  \caption{OOD complete-episode success (\%). ALF/Sci/Web denote
  ALFWorld/ScienceWorld/WebShop; Avg. equally weights environment means
  before rounding for display.
  All entries are means over three independent training runs.
  Qwen3-8B environment entries additionally report SD.
  Full environment-level results for Qwen3-14B and Llama appear in
  Table~\ref{tab:crossmodel_ood_full}.
  Bold marks column maxima.}
  \label{tab:mitigation_main}
  \label{tab:crossmodel_objectives}
\end{table*}

\subsection{A Possible Training Explanation}
\label{sec:optimization_explanation}

SFT minimizes the prediction loss of recorded responses. Consider a
recorded purchase action: a model may predict it by checking the selected
product against the task requirements, or by continuing a familiar
sequence of option selections. Either route can reduce the prediction
loss on the same demonstrated response. Fitting these demonstrations alone does not ensure that
the model will adjust its action when the task changes.

\section{Mitigating Task Insensitivity with Task-Perturbed NLL}
\label{sec:method}
\label{sec:tpnll_objective}

We introduce \textbf{Task-Perturbed Negative Log-Likelihood (TPNLL)} to
strengthen the dependence of response prediction on task information.
Its auxiliary loss discourages the contrast in prediction costs under
the original and replacement tasks from weakening relative to initialization.

For an SFT example $(x_i,y_i)$, we write $x_i=(g_i,c_i)$, where $g_i$
is the task and $c_i$ is the remaining context.
We replace $g_i$ with a distinct task among the ten most lexically
similar tasks in the training pool, yielding
$\tilde{x}_i=(g'_i,c_i)$. We keep $c_i$ and the complete
reasoning--action response $y_i$ fixed across the two task conditions.

Then, let $L_{\mathrm{src}}^{(i)}$ and $L_{\mathrm{rep}}^{(i)}$ be the mean NLL over
all tokens of $y_i$ under $x_i$ and $\tilde{x}_i$, respectively. A frozen copy
of the model at the start of fine-tuning supplies
$L_{\mathrm{src},0}^{(i)}$ and $L_{\mathrm{rep},0}^{(i)}$.
With $\epsilon=10^{-5}$, the current and initialization ratios are
\begin{equation}
\rho_{\mathrm{cur}}^{(i)}=
\frac{L_{\mathrm{rep}}^{(i)}}
     {L_{\mathrm{src}}^{(i)}+\epsilon},
\qquad
\rho_{0}^{(i)}=
\frac{L_{\mathrm{rep},0}^{(i)}}
     {L_{\mathrm{src},0}^{(i)}+\epsilon}.
\label{eq:tpnll_ratios}
\end{equation}
For the same recorded response, a higher ratio means a higher prediction
cost under the replacement task relative to the original task.
Each example uses its own NLL ratio at initialization as a reference.
This accounts for differences between examples at the start of training.
TPNLL penalizes a current ratio below this reference:
\begin{equation}
\mathcal L=
\mathcal L_{\mathrm{SFT}}+
\frac{\lambda}{B}\sum_{i=1}^{B}
\max\!\left(0,
\log\rho_{0}^{(i)}-
\log\rho_{\mathrm{cur}}^{(i)}\right),
\label{eq:tpnll_main}
\end{equation}
where $B$ is the minibatch size and $\lambda=0.3$. The auxiliary term is
inactive when the current ratio meets or exceeds its frozen reference.
Training replacements come only from the training task pool. Evaluation
replacement goals in Section~\ref{sec:explicit_replacement} are excluded
from that pool.

\begin{figure*}[!tbp]
  \centering
  \includegraphics[width=\textwidth,trim=0 5bp 0 17bp,clip]{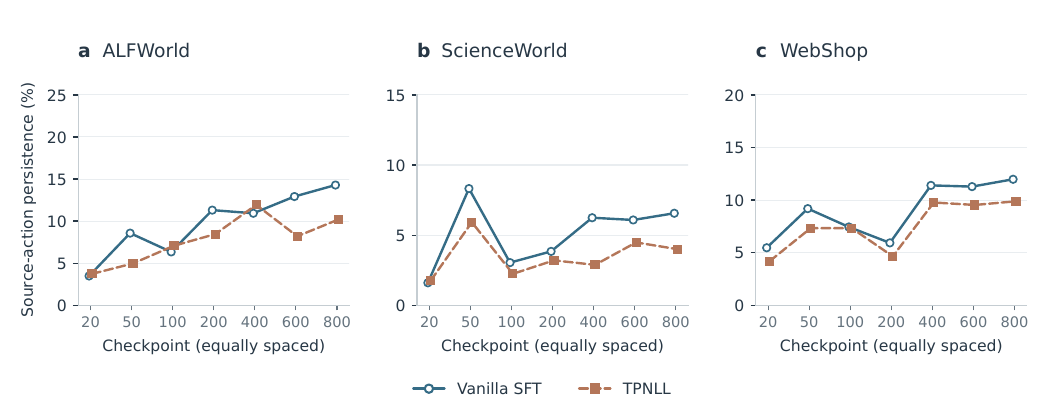}
  \caption{Source-action persistence over all sampled responses
  ($A_{\mathrm{all}}$, \%) under complete task replacement for
  Qwen3-8B trained with Vanilla SFT or TPNLL. Both methods use all five
  sampled responses per task pair as the denominator.
  Panels use separate vertical scales.}
  \label{fig:tpnll_unseen_replacement}
\end{figure*}

\section{Evaluating TPNLL}
\label{sec:experiments}

We evaluate whether TPNLL increases attention to task information,
improves OOD task success, and reduces action persistence after task
replacement. We compare training objectives across model scales and
families (Table~\ref{tab:mitigation_main}), test persistence after task
replacement, and examine how
replacement-task construction and the regularization weight affect the
success gains.
Appendix~\ref{app:rl_evaluation} also evaluates a task-perturbed RL objective
with Qwen3-4B.

\subsection{Setup and Comparisons}

We measure complete-episode success in ALFWorld, ScienceWorld, and
WebShop, allowing at most 30 interaction steps per episode. OOD evaluation
uses task types or product categories held out from training; the ID and
OOD pools are disjoint (Appendix~\ref{app:dataset_split}). Each
model--objective comparison uses three independent training runs with
shared prompts and interfaces, a global batch size of 8, and 800 optimizer
updates. Additional evaluation settings
appear in Appendix~\ref{app:supplementary_protocol}.

We compare five objectives. Vanilla SFT fits the source response. NLL
Difference smoothly favors a higher replacement-task NLL; Margin Hinge
requires an absolute NLL gap of 0.5; Unlikelihood
\citep{welleck2020unlikelihood} suppresses the original
response tokens under the replacement task; and TPNLL uses the
initialization-calibrated NLL ratio. More details are shown in Appendix~\ref{app:intervention}.

\subsection{Does TPNLL Improve Task Success?}
\label{sec:ood_success}

TPNLL achieves the highest cross-environment mean OOD success among
the five objectives on Qwen3-8B, Qwen3-14B, and Llama-3.1-8B-Instruct
(Table~\ref{tab:mitigation_main}). For Qwen3-8B, the average gain over
Vanilla SFT is 5.0 percentage points, with improvements in every environment.

\subsection{Task Attention under TPNLL}
\label{sec:tpnll_attention_eval}

\begin{figure*}[!tb]
  \centering
  \includegraphics[width=\textwidth]
    {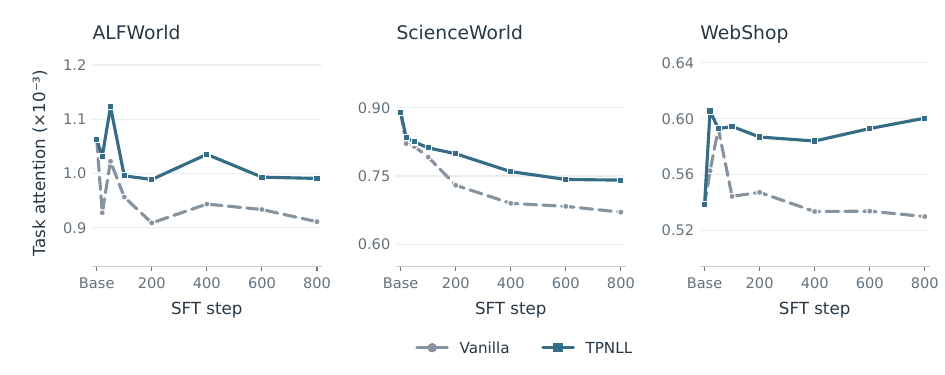}
  \caption{Mean per-task-token attention density ($10^{-3}$) for Qwen3-8B
  under Vanilla SFT and TPNLL. Both methods share Base; panels use
  separate vertical scales.}
  \label{fig:tpnll_attention}
\end{figure*}

TPNLL yields higher task-token attention than Vanilla at the matched SFT
checkpoints (Figure~\ref{fig:tpnll_attention}). The Qwen3-8B comparison uses
the fixed-input measurement protocol from Section~\ref{sec:attention};
shared evaluation settings appear in Appendix~\ref{app:tpnll_attention}.
Using the same prompt--response pairs controls for changes in response
content and length, so the comparison reflects attention allocation under
the two objectives.

\subsection{Does TPNLL Reduce Action Persistence?}
\label{sec:tpnll_replacement_eval}

TPNLL reduces source-action persistence relative to Vanilla SFT on the
complete replacement probes (Figure~\ref{fig:tpnll_unseen_replacement}).
Both Qwen3-8B conditions follow the shared checkpoint schedule and
sampling protocol in Appendix~\ref{app:replacement_scoring}, through
the same step-800 endpoint.
These probes complement complete-episode success by testing whether the
model continues to reuse a recorded action after the task requirement
makes it unsuitable. Replacement goals are excluded from the training
task pool, so the comparison also tests transfer beyond the training
contrasts.

\subsection{Effect of Replacement-Task Construction}
\label{sec:replacement_construction}

TPNLL obtains its training contrast by changing the task. To test the
effect of replacement choice, we compare the default lexically similar
but distinct tasks with random replacements for Qwen3-8B, using Vanilla
SFT as a reference
(Appendix~\ref{app:replacement_quality}).

\begin{table}[!htbp]
  \centering
  \small
  \setlength{\tabcolsep}{4pt}
  \begin{tabular}{lccc}
    \toprule
    Training condition & ALF & Sci & Web\\
    \midrule
    Vanilla & 54.9 & 25.4 & 47.3\\
    TPNLL: similar task & 59.4 & 32.3 & 51.0\\
    TPNLL: random task & 55.5 & 24.7 & 48.0\\
    \bottomrule
  \end{tabular}
  \caption{Qwen3-8B OOD success (\%) by replacement-task construction.
  All entries report means over three runs.}
  \label{tab:replacement_quality}
\end{table}

Random replacement tasks yield success rates close to Vanilla,
whereas the default replacements give larger gains
(Table~\ref{tab:replacement_quality}). Thus, the construction of the
task contrast matters for the observed benefits. One possible
explanation is that random sampling can yield unrelated goals under which
the retained response is already unlikely, weakening the training signal.

\subsection{Regularization-Weight Ablation}
\label{sec:sft_lambda_ablation}
\label{app:sft_lambda_ablation}

We vary $\lambda$ while holding the model, data, training budget, and
evaluation settings fixed. The $0$ and $0.3$ rows repeat the Vanilla and
default TPNLL results for comparison. Table~\ref{tab:sft_lambda_ablation}
shows that the best mean varies by environment: $0.7$ is best in ALFWorld,
$0.3$ in ScienceWorld, and $0.3$ and $0.5$ tie in WebShop. Increasing the
weight therefore does not consistently improve success; we retain
$\lambda=0.3$ for the main experiments.

\begin{table}[htbp]
  \centering
  \small
  \setlength{\tabcolsep}{4pt}
  \begin{tabular}{cccc}
    \toprule
    $\lambda$ & ALF & Sci & Web\\
    \midrule
    $0$   & $54.9{\pm}4.4$ & $25.4{\pm}0.7$ & $47.3{\pm}2.8$\\
    $0.3$ & $59.4{\pm}2.8$ & $\mathbf{32.3{\pm}0.5}$
      & $\mathbf{51.0{\pm}2.5}$\\
    $0.5$ & $59.4{\pm}6.0$ & $30.5{\pm}3.3$ & $\mathbf{51.0{\pm}1.6}$\\
    $0.7$ & $\mathbf{61.3{\pm}3.1}$ & $28.4{\pm}3.9$ & $46.9{\pm}3.2$\\
    \bottomrule
  \end{tabular}
  \caption{Qwen3-8B OOD success (\%) by TPNLL weight $\lambda$:
  mean $\pm$ SD over three runs. Bold marks the highest displayed means.}
  \label{tab:sft_lambda_ablation}
\end{table}

\section{Conclusion}

Language agents can follow contextual cues without usable task descriptions,
even when they acknowledge missing requirements. During SFT, task attention
declines while unsuitable actions increasingly persist after task replacement
in many settings. TPNLL strengthens task dependence and improves OOD success.
These findings distinguish coherent continuation from goal-directed action,
motivating training objectives and evaluations that test whether task
requirements constrain decisions.

\section*{Limitations}
Our study mainly focuses on supervised fine-tuning, and we conduct RL
experiments only with Qwen3-4B due to limited computational resources.
We do not fully explain why training makes models rely more on observations
than on task descriptions; we leave this question to future work.

\section*{LLM usage}
We used generative AI tools to assist with literature searches, discussion
and refinement of research hypotheses, code implementation, and data analysis.
We also used these tools to revise, polish, and review the
manuscript. The authors reviewed and verified the AI-assisted work.

\bibliography{refs}

\clearpage
\appendix

\renewcommand{\floatpagefraction}{0.90}
\renewcommand{\dblfloatpagefraction}{0.90}

\section{Existing-Model Evaluation}
\label{app:existing_agent_evaluation}

\subsection{Decision States and Evaluation Protocol}
\label{app:existing_state_selection}
\label{app:response_classification}

\paragraph{States and inputs.}
Table~\ref{tab:placeholder_behavior} and Figure~\ref{fig:context_next_step_groups}
use the same 400 ongoing decision states: 150 each from ALFWorld and
ScienceWorld, and 100 from WebShop. Selection in the first two environments
balances operation stages and source tasks, removes duplicate full prompts,
and retains at most three states per source task and two per normalized
current state. WebShop uses 100 distinct source tasks. All explicit task
copies become \texttt{xxxx}; history, observation, action menus, and operating
instructions remain unchanged (Appendix~\ref{app:native_agent_prompts}).

\paragraph{Goal-disclosure checks.}
Deterministic scripts extract the full source task from the native prompt
template and replace every exact occurrence with \texttt{xxxx}, including
copies embedded in earlier observations. Validation requires the task
field to equal \texttt{xxxx} and checks that the full source-task string
does not remain anywhere in the prompt, ignoring case; the WebShop check
also normalizes whitespace. WebShop additionally excludes histories with
visible search queries, which may repeat requested product attributes.
These checks use the input text, without an LLM judge. Object names,
completed operations, and selected options remain visible as operational
context; the check targets explicit task disclosure rather than removing
every possible clue about the goal.

\paragraph{Generation.}
Each model generates ten responses per state. GPT-5.6-Sol
(\texttt{gpt-5.6-sol-2026-07-09}) uses temperature 1, top-$p$ 1,
and high reasoning effort. DeepSeek-V4.1-Flash
(\texttt{ali-deepseek-v4.1-flash}), Qwen3.5-Plus, and Qwen3-8B/32B
use temperature 0.7 and top-$p$ 0.95. Output budgets are 8,192 tokens for
the three API models and 2,048 for Qwen3-8B/32B. DeepSeek uses its default
reasoning mode; Qwen3.5-Plus has native reasoning disabled, while
Qwen3-8B/32B have native thinking enabled. All models receive the same
analysis-output instruction:
\begin{lstlisting}[style=promptstyle]
Choose one next response from the current state, output your thinking steps in <think></think>, and your final answer in
<action>one admissible environment action</action>
or if you want to ask clarification questions, you can answer with
<inquiry>one question for the user</inquiry>
\end{lstlisting}
This yields 4,000 responses per model and 20,000 across the five models.
The instruction requests analysis but does not specifically ask models
to check or restate the task. Complete returned text and separate API
reasoning fields are saved. The same final-response parsing is used for
Table~\ref{tab:placeholder_behavior} and Figure~\ref{fig:context_next_step_groups};
details appear in Appendix~\ref{app:task_checking}.
Inquiry combines explicit \texttt{<inquiry>} outputs and ScienceWorld's
native \texttt{task} command, which also requests task information;
the latter is excluded from Act and counted as Inquiry.

\input{task_checking_appendix}

\subsection{Context-Guided Continuation Groups}
\label{app:context_continuation}

Figure~\ref{fig:context_next_step_groups} compares states where recorded
operations suggest a specific follow-up (cued continuation) with states
where an object, destination, or procedure remains to be selected
(unresolved choice). These groups describe local operational progress
under \texttt{xxxx}; they do not establish the user's goal.

\paragraph{Grouping procedure.}
The grouping uses the visible history, current observation, and action
menu. It proceeds in three stages:
\begin{enumerate}
  \item \textbf{Recover operational progress.} Pair each historical
  action with the following observation and use the feedback to confirm
  completed operations. Track the relevant objects, locations, and
  apparatus states, clearing records when later actions invalidate them.
  \item \textbf{Find supported follow-ups.} Apply the environment-specific
  rules in Tables~\ref{tab:continuation_group_rules}
  and~\ref{tab:science_continuation_rules} to identify the operational
  stage and candidate next actions. In ALFWorld and WebShop, every
  candidate must also appear in the current action menu.
  \item \textbf{Assign a group.} A nonempty candidate set gives
  \emph{cued continuation}. A remaining choice of object, destination,
  attribute, or procedure gives \emph{unresolved choice}. Competing cues,
  failed operations, or unclear stages may instead give \emph{boundary},
  excluded from Figure~\ref{fig:context_next_step_groups}.
\end{enumerate}
Candidates describe local operations, not verified actions for the hidden
goal. Neither the hidden task nor evaluated model responses enter these
rules. The review adjustments are described below.

\paragraph{Worked example.}
For example, confirmed \texttt{take cellphone 1 from desk 1} followed by
arrival at a closed safe supports \texttt{open safe 1} when that command is
available; pickup without a destination leaves the next destination unresolved.

\paragraph{Environment-specific rules.}
The tables give the ordered rules for each environment. Confirmation
checks, event windows, and command matching are specified afterward.

\begin{table*}[!tbp]
  \centering\footnotesize
  \setlength{\tabcolsep}{4pt}
  \renewcommand{\arraystretch}{1.14}
  \begin{tabular}{@{}p{0.17\textwidth}p{0.46\textwidth}p{0.33\textwidth}@{}}
    \toprule
    \raggedright Environment / stage & Trigger and state condition & Candidate actions / assignment\\
    \midrule
    \raggedright ALFWorld: completed treatment &
    The latest confirmed treatment is the last event, or it concerns the
    held object and occurred after the most recent confirmed arrival. &
    No rule-generated candidate; unresolved unless a review adjustment applies.\\
    \raggedright ALFWorld: treatment available &
    Confirmed pickup of held \texttt{X}, followed by a confirmed arrival.
    The menu offers treatment of \texttt{X} not already recorded as completed. &
    All such \texttt{clean X with Z}, \texttt{heat X with Z}, or
    \texttt{cool X with Z} commands; cued. Treatment takes priority over placement.\\
    \raggedright ALFWorld: placement available &
    Confirmed pickup followed by arrival at \texttt{D}; \texttt{D}
    differs from the pickup origin. No treatment rule applies. &
    \texttt{move X to D}, if present in the menu; cued.\\
    \raggedright ALFWorld: closed receptacle &
    The same pickup and arrival conditions hold, but neither treatment
    nor placement supplies a candidate. &
    \texttt{open D}, if present in the menu; cued.\\
    \midrule
    \raggedright WebShop: attribute coverage &
    Split the current page at \texttt{[SEP]}. A non-option token immediately
    preceding an option value identifies its attribute field; values
    are drawn from the option-click menu. Each historical
    \texttt{click[value]} must map to exactly one current field. Its
    preceding and following pages must share the current product title
    and contain that value. &
    \texttt{click[buy now]} when every visible field has a verified
    selection; cued. Any uncovered field gives unresolved choice.
    Ambiguous, unverified, repeated, or revised selections are excluded.\\
    \bottomrule
  \end{tabular}
  \caption{ALFWorld and WebShop continuation rules. ALFWorld uses the first
  matching stage in the listed order. Every candidate must occur in the
  current action menu. States with no qualifying continuation remain
  unresolved before the existing review adjustments.}
  \label{tab:continuation_group_rules}
\end{table*}

\begin{table*}[!tbp]
  \centering\footnotesize
  \setlength{\tabcolsep}{4pt}
  \renewcommand{\arraystretch}{1.14}
  \begin{tabular}{@{}p{0.17\textwidth}p{0.46\textwidth}p{0.33\textwidth}@{}}
    \toprule
    \raggedright ScienceWorld stage & Trigger and state condition & Candidate actions / assignment\\
    \midrule
    \raggedright Linked circuit &
    The last event is a confirmed connection. Check the last eight
    events for a confirmed connection with different normalized feedback
    and a shared component.
    \texttt{disconnect}, pickup, put-down, or navigation clears the
    connection sequence. &
    Linked connections give boundary: the next endpoint or operation
    stage is unclear. A single confirmed connection gives unresolved
    choice with no candidate.\\
    \raggedright Plane: timer running &
    Confirmed placement at \texttt{P}, with $t-i_{\mathrm{place}}\leq6$,
    and the latest confirmed stopwatch toggle is activation. Any
    subsequent pickup, put-down, other move, or navigation clears placement. &
    \texttt{wait}, \texttt{wait1}, \texttt{deactivate stopwatch}, and
    \texttt{look at O} for \texttt{O} in
    \{\texttt{X}, \texttt{Y}, \texttt{P}, \texttt{stopwatch}\}; cued.\\
    \raggedright Plane: just placed &
    Placement is the last event and the timer-running rule does not
    apply. Some preceding action must mention \texttt{stopwatch}. &
    \texttt{activate stopwatch}; cued. Without the preceding mention,
    assign boundary.\\
    \raggedright Plane: timer stopped &
    Placement remains valid within the same window; the last action is
    a confirmed \texttt{deactivate stopwatch} at or after placement. &
    \texttt{look at O} for the same object set as above; cued.\\
    \raggedright Stopwatch only &
    The last action mentions \texttt{stopwatch}, or starts with
    \texttt{wait} and one of the last four actions mentions it.
    At least one confirmed stopwatch toggle exists; no earlier plane
    stage applies. &
    \texttt{look at stopwatch}; additionally \texttt{wait},
    \texttt{wait1}, and \texttt{deactivate stopwatch} if the latest
    confirmed toggle is activation. Cued, with a timer-only subtype.\\
    \raggedright Heating &
    Confirmed loading of \texttt{X} into \texttt{D}, whose latest
    confirmed toggle is activation, with $t-i_{\mathrm{load}}\leq4$.
    Any subsequent pickup, put-down, other
    move, or navigation clears loading. The last action starts with
    \texttt{move}, \texttt{activate}, \texttt{close}, \texttt{wait},
    \texttt{use thermometer}, or \texttt{look at}. &
    \texttt{wait}, \texttt{wait1}, \texttt{look at X}; add
    \texttt{use thermometer on X} if the menu mentions a thermometer.
    For an oven, add \texttt{close oven} unless it was the last action.
    Cued.\\
    \raggedright Mixing &
    In the last five events, at least two distinct ingredients were
    successfully poured into the same \texttt{B}, with a pour as the
    last event. \texttt{mix}, \texttt{dunk}, or navigation clears the
    pouring sequence. &
    \texttt{mix B}; cued. A single ingredient leaves the next ingredient
    or procedure unresolved.\\
    \bottomrule
  \end{tabular}
  \caption{ScienceWorld continuation rules, evaluated in the listed
  order. Navigation comprises \texttt{go} and \texttt{teleport}; pickup
  and put-down comprise \texttt{pick up} and \texttt{put down}. Commands
  instantiate recorded object and apparatus names. Unlike ALFWorld and
  WebShop, these template candidates need not appear as fully instantiated
  commands in the menu. Other stages supply no rule-generated candidate.}
  \label{tab:science_continuation_rules}
\end{table*}

\paragraph{Event and matching conventions.}
We index the visible historical actions from $0$ to $t$ and pair each
action with the next observation; the current observation supplies the
feedback for action $t$. Strings are case-folded and consecutive
whitespace is collapsed. Command templates in the tables match the whole action;
feedback matches are exact unless explicitly described as prefix or
suffix checks. A record at index $i$ is within the placement or heating
window when $t-i\leq6$ or $t-i\leq4$, respectively. In contrast, the mixing
window contains the last up to five events, including $t$.

For ALFWorld, \texttt{take X from R} is confirmed by the exact feedback
\texttt{you pick up the X from the R.}; \texttt{go to D} requires the
prefix \texttt{you arrive at D.}; and \texttt{V X with Z}, where
\texttt{V} is \texttt{clean}, \texttt{heat}, or \texttt{cool}, requires
the prefix \texttt{you V the X }. Confirmed pickup stores the held object
and its origin, arrival stores the current receptacle, and treatment
records the operation--object pair as completed. A placement command
clears the held-object record when feedback begins with
\texttt{you move } or is missing. Missing pickup or arrival feedback
clears the corresponding record; nonempty feedback that fails its
confirmation check does not update that record.

For ScienceWorld, apparatus toggles require exact feedback
\texttt{the Z is now activated.} or
\texttt{the Z is now deactivated.}, with \texttt{Z} bound to the
stopwatch, stove, or oven. Plane placement requires
\texttt{move X to P}, where \texttt{P} begins with
\texttt{inclined plane}, and exact feedback
\texttt{you move the Y to the inclined plane.}; both the command name
\texttt{X} and returned name \texttt{Y} are retained for inspection.
Loading a stove or oven requires feedback beginning with
\texttt{you move } and ending with \texttt{to the D.}.
\texttt{pour X in B} or \texttt{pour X into B} requires feedback beginning
with \texttt{you pour } and ending with \texttt{into the B.}.
The reset conditions in Table~\ref{tab:science_continuation_rules} apply
to intervening commands, even if those commands do not confirm a
successful operation.

For the circuit check, confirmed feedback has the form
\texttt{U is now connected to V}, with an optional final period.
Component names are extracted after \texttt{terminal N on},
\texttt{anode on}, or \texttt{cathode on}; connections with unparseable
endpoints are not accepted. WebShop's product title is the token preceding
\texttt{Price:}. Attribute extraction carries the latest field name
forward across its option-value tokens. A value assigned to multiple
fields is ambiguous; multiple verified clicks for one field count as a
revision and are excluded from the confirmation cohort.

\paragraph{Implementation and review.}
The following pseudocode summarizes grouping from visible input:
\begin{lstlisting}[style=promptstyle,caption={Continuation grouping from visible input.},label={lst:continuation_grouping}]
Input: history H, current observation O, menu M
Output: group, supporting events, candidates C

Normalize case and whitespace in H, O, and M.
Pair each action with the following observation;
use O as feedback for the final historical action.
Replay events with the confirmation/reset checks.
Apply the ordered environment rules to obtain
the stage, supporting events, and candidates C.
For ALFWorld/WebShop, require every candidate in M.
Set group = cued if C is nonempty, else unresolved;
retain boundary assignments made by stage rules.
Apply the existing review adjustments described below.
Deduplicate and normalize the final candidates.
Return group, supporting events, and candidates.
\end{lstlisting}

Scripts implement these checks without an LLM judge. ALFWorld and
ScienceWorld use exploratory groupings with post-hoc human review;
WebShop labels were frozen before generation. Each final label records
supporting events, candidates, and its rule or review reason. Hidden
tasks, future recorded actions, and individual model outputs are excluded.

Final ALFWorld counts are 104 cued, 35 unresolved, and 11 boundary;
ScienceWorld counts are 50, 57, and 43. Eleven ScienceWorld cases with confirmed
stopwatch operation but unconfirmed apparatus form a timer-only subtype.
Unclear circuit endpoints or stages and ambiguous recovery sequences are boundary.

\paragraph{WebShop selection.}
\label{app:webshop_confirmation}
All cases show a product page, two history steps, and an available
purchase action. Selecting every visible attribute supplies the cued
case; an outstanding attribute supplies the unresolved case.
We exclude search queries, errors, repeated or revised clicks,
unverified selections, ambiguous attribute mappings, and pages without
options. Fixed-hash ordering alternates selection between groups, with
one state per task and at most one product title per group. Excluding
tasks and normalized prompts from earlier cohorts yields 50 states per
group. Recorded next actions and episode success do not determine inclusion.

\paragraph{Group metrics.}
Table~\ref{tab:continuation_group_rates} gives the counts behind
Figures~\ref{fig:context_next_step_groups} and~\ref{fig:context_task_acknowledgment}:
action rates use all ten responses per state; acknowledgment rates use
ordinary actions and the labels in Appendix~\ref{app:task_checking}.
Boundary states remain in
Table~\ref{tab:placeholder_behavior}. Pointwise 95\% percentile intervals
use 5,000 bootstrap resamples of states within each model, environment, and
group, keeping all ten responses together. States without ordinary actions
remain eligible; an acknowledgment rate is omitted only when a resampled
group has no ordinary actions.

\begin{table*}[!tbp]
  \centering\small
  \setlength{\tabcolsep}{3.5pt}
  \begin{tabular}{@{}llrrrr@{}}
    \toprule
    & & \multicolumn{2}{c}{Cued continuation} & \multicolumn{2}{c}{Unresolved choice}\\
    \cmidrule(lr){3-4}\cmidrule(lr){5-6}
    Model & Environment & Actions / all & Explicit / actions (\%)
    & Actions / all & Explicit / actions (\%)\\
    \midrule
    \ContextGroupRatesRows
    \bottomrule
  \end{tabular}
  \caption{Action and explicit-acknowledgment counts for the continuation
  groups. Explicit denotes a semantic acknowledgment of missing task
  requirements, rather than a literal mention of \texttt{xxxx}.}
  \label{tab:continuation_group_rates}
\end{table*}

\begin{table*}[!tbp]
  \centering\small
  \setlength{\tabcolsep}{5pt}
  \begin{tabular}{@{}llccccc@{}}
    \toprule
    & & \multicolumn{3}{c}{Episode success (\%)} & \multicolumn{2}{c}{$C^{+}$ behavior (\%)}\\
    \cmidrule(lr){3-5}\cmidrule(lr){6-7}
    Model & Environment & $F$ & $C^{+}$ & $C^{-}$ & No query & Received\\
    \midrule
    \RolloutPairedRows
    \bottomrule
  \end{tabular}
  \caption{Rollout outcomes (\%; 150 episodes per cell). $F$: full task;
  $C^{+}/C^{-}$: corrupted task with/without selective clarification.
  No query: neither inquiry nor native task query; Received: original
  task disclosed.}
  \label{tab:selective_rollout}
\end{table*}

\subsection{Rollouts with Selective Task Clarification}
\label{app:selective_rollout}

This experiment asks whether agents obtain missing task requirements
before completing an episode. It uses partial or conflicting descriptions
rather than \texttt{xxxx}, and starts from native initial states.

\paragraph{Conditions.}
The five models each run 150 episodes per environment under three
conditions: full task ($F$), corrupted task with clarification ($C^{+}$),
and the same corrupted task without clarification ($C^{-}$).
Matched episodes start from the same initial state and are scored
against the original goal, giving 6,750 episodes in total.

ALFWorld corruption shortens descriptions to keywords, often retaining
the object, destination, and required state. ScienceWorld removes a
result-to-box rule or comparison direction. WebShop introduces conflicting
attribute requirements. All explicit task copies in a prompt receive the
same change. Table~\ref{tab:corruption_examples} gives abbreviated examples
of these operators from the earlier task-corruption evaluation; these
illustrate the transformations rather than reproduce full rollout prompts.

\begin{table*}[!tbp]
  \centering\small
  \setlength{\tabcolsep}{5pt}
  \renewcommand{\arraystretch}{1.12}
  \begin{tabular}{@{}lp{0.37\textwidth}p{0.42\textwidth}@{}}
    \toprule
    Environment & Original requirement & Corrupted description\\
    \midrule
    ALFWorld & put a kettle in the shelf & kettle shelf\\
    ScienceWorld & Above 150$^\circ$C: red box; below: green box
      & Above 150$^\circ$C: the box; below: the box\\
    WebShop & size: 5ft 3in $\times$ 7ft 5in
      & size: 5ft 3in $\times$ 7ft 5in and size: 7ft 8in\\
    \bottomrule
  \end{tabular}
  \caption{Abbreviated corruption examples. The edits remove an operation
  relation, leave an answer mapping unspecified, or introduce conflicting
  attribute values.}
  \label{tab:corruption_examples}
\end{table*}

\paragraph{Clarification.}
In $F$ and $C^{+}$, agents may ask a question through \texttt{<inquiry>}
or ScienceWorld's \texttt{task} command. GPT-5.4 judges whether the question
concerns a missing goal requirement that the original task would resolve.
If so, the agent receives the full original task; otherwise it receives
an empty reply. Questions about object locations or execution procedures
do not trigger disclosure. Both query interfaces are blocked in $C^{-}$.
Appendix~\ref{app:clarification_judge_prompt} summarizes the judge inputs
and decision criteria.

\paragraph{Results.}
In $C^{+}$, 1,849/2,250 episodes (82.2\%) contain no query
(Table~\ref{tab:selective_rollout}). For GPT-5.6-Sol in ScienceWorld,
59 of the 62 paired episodes that succeed under $F$ but fail under
$C^{+}$ contain no query. Inspected failures include choosing a different
box from the full-task run without asking for the missing sorting rule.
In these failures, the agent reaches its final choice without asking for
the missing task requirement.

\section{Task-Replacement Evaluation}
\label{app:task_replacement}

We replace a complete task while holding the non-task context fixed,
then measure how often the model retains the recorded source action.
Replacement tasks are absent from the original seven-model SFT data.

\subsection{Task Construction and Validation}
\label{app:replacement_protocol}
\label{app:unseen_id_replacement}

We draw replacement goals from native inventories, excluding task texts
and structured goals present in SFT. Deterministic goal checks and native
execution verify that the non-task context is unchanged, the source action
violates a new requirement, and the replacement goal remains feasible.
Reversible substitution checks every explicit task copy. Independent
replays complete the source goal beginning with the recorded action and
the replacement goal; shared steps suitable for both are excluded.

\paragraph{ALFWorld.}
We extract task type, object, destination, and processing requirements
from native metadata and PDDL parameters, then match the source command
to \texttt{take} or \texttt{move}. Object-selection replacements require a
different object type with an available take action. Placement replacements
retain object type but change destination or cleaning/heating/cooling
requirements; native PDDL evaluation must show that the old placement
immediately completes the source goal but fails the new goal.
Replay changes only task text and PDDL goal, preserving the initial physical
state. Observations, action menus, and physical facts must match along
the history. A native planner or direct navigation-and-placement route
must complete the new goal in a fresh replay. A next-action mismatch may
still be repairable later.

\paragraph{ScienceWorld.}
Task-template parsers extract object/category, destination, threshold,
and answer mapping; the native action parser binds commands to object IDs
and checks legality. Object types determine classification targets;
material properties and task metadata determine conductivity/melting-point
branches; friction uses native target-surface metadata. Retained source
actions select an object outside the required set, a conflicting destination,
or an incompatible answer. Answer-rule changes exclude shared measurements.

Replay restores the history before activating the new goal, without world
modifiers or time advancement; goals requiring earlier progress instead
replay the history under the replacement. Object trees and legal menus
must match, with unchanged nonthermal properties and at most 0.1$^\circ$C
temperature difference per object. Separate replays must complete both goals.

\paragraph{WebShop.}
Native goal dictionaries must match in product ID, category, price limit,
other attributes, and option-field names; exactly one option value changes.
A selection-state source click must choose the old value, with the new value
available; a purchase-state configuration must conflict with that value.
Replayed pages, selected options, and physical session state must match.
The source continuation must earn full reward under the source goal and
less under the replacement; correcting the option before purchase must
restore full reward. Selection-state suffixes leave that option fixed,
and all continuations fit the remaining budget. Option errors can be
corrected before purchase, which commits the configuration.

\begin{table*}[!tbp]
  \centering\footnotesize
  \renewcommand{\arraystretch}{0.92}
  \setlength{\tabcolsep}{3pt}
  \begin{tabular}{@{}llccccc@{}}
    \toprule
    Model & Env. & Base & Step 20 & Step 800
    & $\Delta_{0\to800}$ (pp) & $\Delta_{20\to800}$ (pp)\\
    \midrule
    \AllResponseIDEndpointRows
    \bottomrule
  \end{tabular}
  \caption{Source-action persistence over all sampled responses
  $A_{\mathrm{all}}$ (\%) under task replacement at
  Base and SFT steps 20 and 800. Changes are in percentage points.}
  \label{tab:replacement_endpoints}
\end{table*}

\subsection{Generation and Scoring}
\label{app:replacement_scoring}

The seven models in Section~\ref{sec:attention} are evaluated at Base and
SFT steps 20, 50, 100, 200, 400, 600, and 800 on 161 ALFWorld, 125
ScienceWorld, and 172 WebShop pairs (458 total). Each distinct source or
replacement input receives five outputs, reused for identical inputs.

The parser excludes complete \texttt{<think>} or \texttt{<thinking>} blocks;
a closing tag without an opening tag excludes preceding text, while an
unclosed block yields no action. Exactly one nonempty final \texttt{<action>}
and no \texttt{<inquiry>} are required. Case and whitespace are ignored.
$A_{\mathrm{all}}$ pools source-action matches over all responses, including
unparsed responses as nonmatches. TPNLL, restatement, mixed-environment,
and RL comparisons share this denominator; the Qwen3-8B diagnosis and
method comparison share their Vanilla outputs.

Table~\ref{tab:replacement_endpoints} reports early and final persistence.
Pointwise 95\% intervals resample source-task clusters
\citep{efron1979bootstrap}, jointly resampling match and response counts
before recomputing their ratio. Endpoint differences use the same cluster
draw at both checkpoints. These intervals concern evaluation cases, not
training seeds; persistence concerns the next action, not eventual success.

\subsection{Task Restatement during SFT}
\label{app:task_restatement}

This experiment asks whether demonstrating task restatement before each
action is sufficient to prevent source-action persistence.

\begin{figure*}[!tbp]
  \centering
  \includegraphics[width=\textwidth]{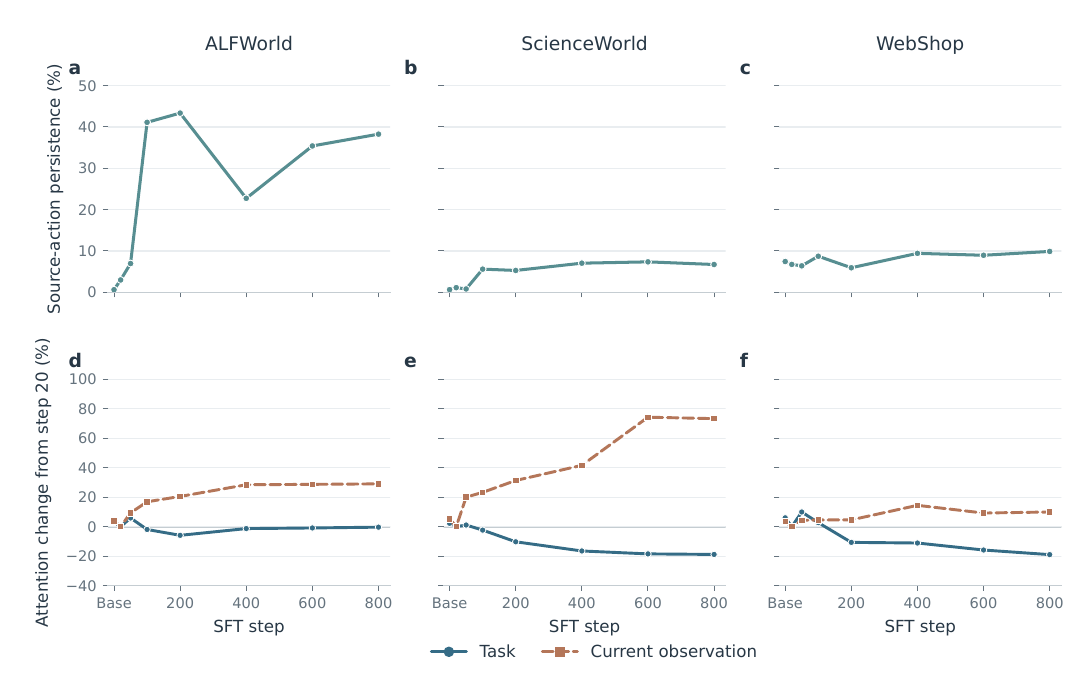}
  \caption{Qwen3-8B trained on task-restating demonstrations.
  Top: source-action persistence.
  Bottom: per-token attention density relative to step 20.
  Base denotes the model before task-restatement SFT.}
  \label{fig:task_restatement_sft}
\end{figure*}

\paragraph{Training.}
We collect GPT-5.4 trajectories using the following final instruction in
place of the native response instruction:
\begin{lstlisting}[style=promptstyle]
Now it's your turn to take one action for the current step.
Your response must contain one <thinking>...</thinking> block, starting with the task restatement, followed by exactly one <action>...</action> block.
\end{lstlisting}
We retain trajectories that pass the native environment success test and
uniformly sample step-level prompt--response rows per environment,
without replacement. For SFT, we restore each row's original native
prompt and retain the GPT-5.4 response unchanged. Qwen3-8B is initialized
independently for each environment and trained for 800 steps with global
batch size 8, learning rate $10^{-5}$, 50 warmup steps, and cosine decay.

\paragraph{Evaluation.}
Base and the seven checkpoints use the same 458 pairs, five outputs per
input, parser, and shared Base reference as
Appendix~\ref{app:replacement_scoring}. Evaluation uses native prompts;
task restatement is requested only during demonstration collection.
Attention follows Appendix~\ref{app:attention} on fixed cases, with changes
relative to step 20.

\paragraph{Findings.}
Source-action persistence increases from step 20 to 800 in the 3
environments, even with task-restating demonstrations
(Figure~\ref{fig:task_restatement_sft}). Attention to the task
description also decreases as training progresses. Thus, task-restating
training demonstrations do not ensure task-sensitive action selection.

\label{app:restatement_case}
Table~\ref{tab:task_restatement_example} shows a step-800 ALFWorld response
that repeats \texttt{put two cellphone in safe.} but selects
\texttt{move cellphone 1 to drawer 1}. Native replay verifies this
seven-action solution: go to and open safe 1, place cellphone 1, go to
shelf 1, take cellphone 2, return to safe 1, and place it inside.
The environment then reports success. The table quotes part of the
saved response and the first three verified actions.

\section{Multi-Environment SFT}
\label{app:mixed_environment_sft}

\begin{figure*}[!tbp]
  \centering
  \includegraphics[width=\textwidth]{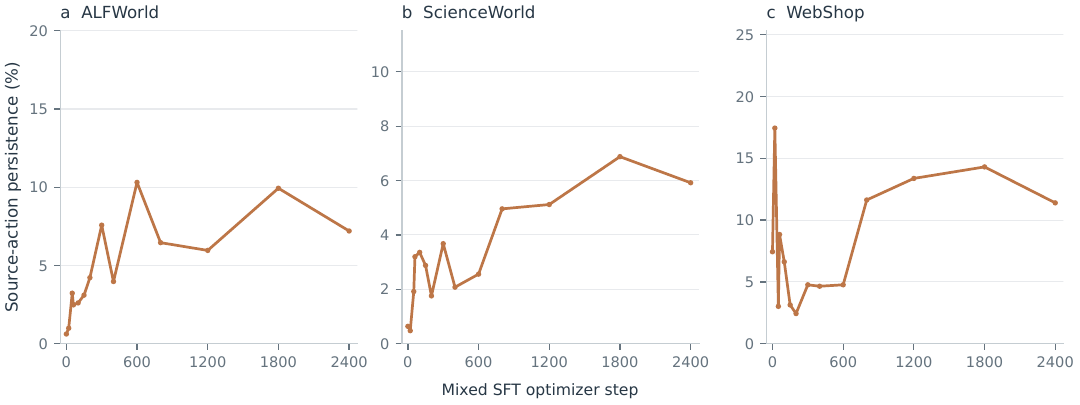}
  \caption{Source-action persistence under complete task replacement during
  mixed-environment SFT of Qwen3-8B. Curves include Base and all 13 evaluated
  checkpoints at their actual optimizer steps.}
  \label{fig:mixed_environment_replacement}
\end{figure*}

\begin{table*}[!tbp]
  \centering\small
  \setlength{\tabcolsep}{4pt}
  \renewcommand{\arraystretch}{1.08}
  \begin{tabular}{@{}lcccc@{}}
    \toprule
    Environment & Base & Mixed-2400
      & \shortstack{Change from Base\\(pp)}
      & \shortstack{Alternative-action\\match (\%)}\\
    \midrule
    \AlignedMixedAlfworldRow
    \AlignedMixedScienceWorldRow
    WebShop & 7.44\% & 11.40\% (98/860)
      & $+3.95$ & 77.79\\
    \bottomrule
  \end{tabular}
  \caption{All-response source-action persistence before and after mixed-environment
  SFT, alongside alternative-action match at Mixed-2400. Counts show
  source-action matches over all sampled responses where available.}
  \label{tab:mixed_replacement}
\end{table*}

\paragraph{Protocol.}
To test broader supervision, Qwen3-8B receives full-parameter Vanilla SFT
on the combined original prompt--response pairs from the three datasets
in Appendix~\ref{app:dataset_split}. Training uses batch size 8, learning
rate $10^{-5}$, cosine decay, and 150 warmup steps. We evaluate Base and
steps 20, 50, 60, 100, 150, 200, 300, 400, 600, 800, 1200, 1800, and 2400.
Replacement uses the same 458 pairs, five responses per distinct input,
and the shared Base reference, parser, and scoring in
Appendix~\ref{app:replacement_scoring}. Attention follows
Appendix~\ref{app:attention} on fixed cases.

\paragraph{Results.}
Mixed-2400 increases source-action persistence relative to Base in all
three environments (Table~\ref{tab:mixed_replacement});
Figure~\ref{fig:mixed_environment_replacement} shows the checkpoint trajectory.
Alternative-action match compares each response with one replay-verified
feasible next action, using all responses as the denominator.
\begin{figure*}[!tbp]
  \centering
  \includegraphics[width=0.85\textwidth]{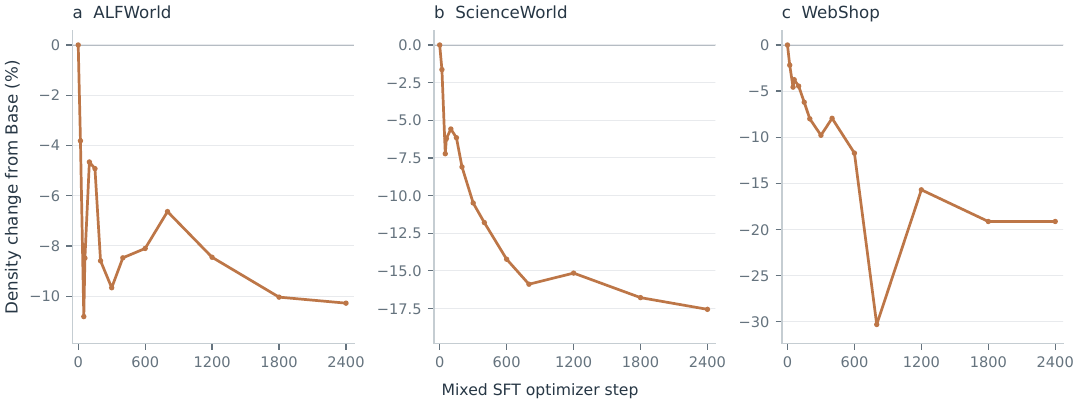}
  \par
  \includegraphics[width=0.85\textwidth]{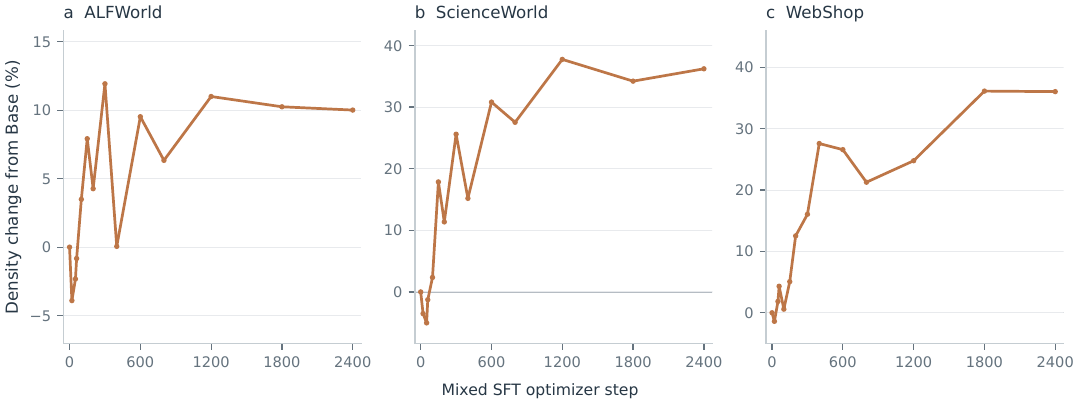}
  \caption{Relative changes in per-token attention density from Base during
  mixed-environment SFT. Top: task; bottom: current observation.}
  \label{fig:mixed_environment_attention}
\end{figure*}

Task attention decreases and current-observation attention increases at
Mixed-2400 in all three environments
(Figure~\ref{fig:mixed_environment_attention}). These changes, together
with action persistence, extend the pattern beyond single-environment SFT.

\section{Attention Analysis}
\label{app:attention}

\subsection{Measurement Protocol}

All seven models in Section~\ref{sec:attention} process the same 500
complete-task prompt--response pairs per environment at each checkpoint.
Responses are teacher-forced, so changes reflect model parameters on fixed
inputs. Equation~\ref{eq:main_task_attention_density} averages attention
from response queries to task keys over layers, heads, and tokens, then
across cases; the same calculation applies to current-observation keys.
Template spans identify each region, combining the original and repeated
task text in ALFWorld. Figure~\ref{fig:response_attention_vanilla} reports
each model's change from Base and their equal-weight mean. Attention and
replacement probes use separate cases and input formats; their association
is descriptive.

\paragraph{TPNLL comparison.}
\label{app:tpnll_attention}
Qwen3-8B Vanilla and TPNLL use this protocol, the same cases, and a shared
initialization reference. Figure~\ref{fig:tpnll_attention} reports raw
per-token densities at initialization and the seven SFT checkpoints.
TPNLL yields higher task-token attention than Vanilla at matched checkpoints.

\subsection{Layer- and Head-Level Changes}
\label{app:attention_granularity}

To locate the decline, we compute task-token density separately for each
layer--head unit on the same 500 cases at initialization and step 800. We count units
whose mean density decreases. For the layer-level result, we first
average heads within each layer, then count decreasing layers.

\begin{table*}[!tbp]
  \centering
  \small
  \setlength{\tabcolsep}{5pt}
  \begin{tabular}{lcccccc}
    \toprule
    & \multicolumn{2}{c}{ALFWorld} & \multicolumn{2}{c}{ScienceWorld}
    & \multicolumn{2}{c}{WebShop}\\
    \cmidrule(lr){2-3}\cmidrule(lr){4-5}\cmidrule(lr){6-7}
    Model & Layers & Layer--heads & Layers & Layer--heads & Layers & Layer--heads\\
    \midrule
    \StepZeroAttentionGranularityRows
    \bottomrule
  \end{tabular}
  \caption{Layers and layer--head units (\%) with decreased mean
  task-token attention from initialization to step 800.}
  \label{tab:attention_granularity}
\end{table*}

Across the 21 model--environment groups, the median fraction with a
decrease is 84.4\% for layers and 76.9\% for layer--head units
(Table~\ref{tab:attention_granularity}), showing that the decline is
distributed across layers and heads.

\section{TPNLL Implementation and Supplementary Results}
\label{app:tpnll_details}

\subsection{Objectives, Data, and Evaluation Settings}
\label{app:intervention}

\paragraph{Training objectives.}
TPNLL uses the ten nearest distinct tasks by edit distance in the same
ID training pool. The non-task context and full reasoning--action response
remain fixed; NLL averages all response tokens. The initialization reference
follows Equation~\ref{eq:tpnll_ratios}.

Let $L_{\mathrm{src}}$ and $L_{\mathrm{rep}}$ be the response NLLs under
the source and replacement tasks. The comparison objectives are
\begin{align}
  \mathcal L_{\mathrm{diff}}
    &=\operatorname{softplus}(L_{\mathrm{src}}-L_{\mathrm{rep}}),\\
  \mathcal L_{\mathrm{margin}}
    &=\max\!\left(0,0.5-(L_{\mathrm{rep}}-L_{\mathrm{src}})\right).
\end{align}
The unlikelihood objective \citep{welleck2020unlikelihood} penalizes the
recorded response tokens under the replacement task:
\begin{equation}
  \mathcal L_{\mathrm{UL}}^{(i)}
  =-\frac{1}{|\mathcal T_i|}\sum_{t\in\mathcal T_i}
  \log\!\left(1-p_\theta(y_{i,t}\mid\tilde{x}_i,y_{i,<t})\right),
\end{equation}
where $\mathcal T_i$ contains all target-response positions. The four
auxiliary objectives use the same task pairs and loss weight 0.3.

\paragraph{Data splits.}
\label{app:dataset_split}
Following \citet{RLVMR}, ALFWorld holds out Cool \& Place and Pick Two
\& Place as OOD types. ScienceWorld holds out the final task type in
each topic; WebShop holds out 30\% of product categories. The remaining
types or categories provide ID training and evaluation tasks. The SFT
sets contain 1,400, 1,525, and 1,500 examples, respectively.

\paragraph{Training and evaluation.}
\label{app:supplementary_protocol}
Each objective uses three independently trained Qwen3-8B, Qwen3-14B, or
Llama-3.1-8B runs: 800 updates, learning rate $10^{-5}$, and batch size 8.
Final-checkpoint evaluation uses 256 episodes per environment, at most
30 steps, temperature 1, and binary environment success. Means and sample
SDs summarize the three runs; environment averages use equal weights
before rounding. Replacement checkpoints follow
Appendix~\ref{app:replacement_scoring}; step counts denote optimizer updates.
Mixed-environment SFT uses the separate 2,400-update budget in
Appendix~\ref{app:mixed_environment_sft}.

\paragraph{Replacement-task construction.}
\label{app:replacement_quality}
The random-replacement condition samples tasks at random in place of the
default nearest-task candidates, while retaining the other settings
(Table~\ref{tab:replacement_quality}).

\subsection{Cross-Model OOD Success}
\label{app:crossmodel_ood_full}

Table~\ref{tab:crossmodel_ood_full} gives the full three-run means and
sample SDs for Qwen3-14B and Llama-3.1-8B-Instruct. Their equally weighted
environment averages appear in Table~\ref{tab:mitigation_main}.

\begin{table*}[!tbp]
  \centering
  \small
  \setlength{\tabcolsep}{4pt}
  \begin{tabular}{@{}lcccccc@{}}
    \toprule
    & \multicolumn{3}{c}{Qwen3-14B} & \multicolumn{3}{c}{Llama-3.1-8B-Instruct}\\
    \cmidrule(lr){2-4}\cmidrule(lr){5-7}
    Method & ALFWorld & ScienceWorld & WebShop & ALFWorld & ScienceWorld & WebShop\\
    \midrule
    Vanilla SFT & $57.9{\pm}1.1$ & $32.4{\pm}2.0$ & $50.9{\pm}5.3$
      & $26.4{\pm}2.2$ & $30.2{\pm}2.9$ & $53.9{\pm}1.0$\\
    NLL Difference & $55.9{\pm}1.8$ & $31.9{\pm}1.0$ & $46.9{\pm}1.0$
      & $25.9{\pm}1.0$ & $29.8{\pm}1.0$ & $55.7{\pm}1.4$\\
    Margin Hinge & $55.6{\pm}0.8$ & $27.7{\pm}2.8$ & $48.3{\pm}1.0$
      & $29.7{\pm}0.8$ & $\mathbf{34.0{\pm}3.1}$ & $53.3{\pm}1.0$\\
    Unlikelihood & $57.2{\pm}1.5$ & $33.2{\pm}0.8$ & $44.1{\pm}1.6$
      & $26.7{\pm}0.5$ & $28.4{\pm}1.8$ & $49.2{\pm}2.3$\\
    \textbf{TPNLL} & $\mathbf{65.1{\pm}2.8}$ & $\mathbf{37.0{\pm}2.0}$ & $\mathbf{55.9{\pm}1.0}$
      & $\mathbf{29.9{\pm}1.6}$ & $33.2{\pm}1.7$ & $\mathbf{57.2{\pm}1.0}$\\
    \bottomrule
  \end{tabular}
  \caption{OOD complete-episode success (\%) for Qwen3-14B and
  Llama-3.1-8B-Instruct: mean $\pm$ sample SD over three runs, with
  256 episodes per environment per run. Bold marks column maxima.}
  \label{tab:crossmodel_ood_full}
\end{table*}

\input{rl_evaluation_appendix}

\input{story_appendix_prompts}

\end{document}

%% file: tables/main_response_rows.tex
\newcommand{\ContextHistoryFullRows}{%
\multicolumn{4}{@{}l}{\textbf{ALFWorld}} \\
GPT-5.6-Sol & 18.7 & 81.3 & 0.0 \\
DeepSeek-V4.1-Flash & 22.5 & 72.8 & 4.7 \\
Qwen3.5-Plus & 10.7 & 88.0 & 1.3 \\
Qwen3-8B & 1.1 & 98.5 & 0.3 \\
Qwen3-32B & 8.1 & 91.5 & 0.3 \\
\midrule
\multicolumn{4}{@{}l}{\textbf{ScienceWorld}} \\
GPT-5.6-Sol & 63.3 & 36.7 & 0.0 \\
DeepSeek-V4.1-Flash & 56.7 & 41.9 & 1.4 \\
Qwen3.5-Plus & 6.1 & 92.3 & 1.7 \\
Qwen3-8B & 3.5 & 91.6 & 4.9 \\
Qwen3-32B & 6.1 & 91.8 & 2.1 \\
\midrule
\multicolumn{4}{@{}l}{\textbf{WebShop}} \\
GPT-5.6-Sol & 5.4 & 94.6 & 0.0 \\
DeepSeek-V4.1-Flash & 3.0 & 96.0 & 1.0 \\
Qwen3.5-Plus & 0.4 & 92.6 & 7.0 \\
Qwen3-8B & 0.0 & 92.1 & 7.9 \\
Qwen3-32B & 0.0 & 99.2 & 0.8 \\
}

%% file: tables/group_rates.tex
\newcommand{\ContextGroupRatesRows}{%
GPT-5.6-Sol & ALFWorld & 1037/1040 & 5/1037 (0.5) & 101/350 & 21/101 (20.8) \\
DeepSeek-V4.1-Flash & ALFWorld & 974/1040 & 887/974 (91.1) & 50/350 & 50/50 (100.0) \\
Qwen3.5-Plus & ALFWorld & 1022/1040 & 828/1022 (81.0) & 204/350 & 201/204 (98.5) \\
Qwen3-8B & ALFWorld & 1035/1040 & 929/1035 (89.8) & 334/350 & 332/334 (99.4) \\
Qwen3-32B & ALFWorld & 1002/1040 & 954/1002 (95.2) & 269/350 & 269/269 (100.0) \\
\midrule
GPT-5.6-Sol & ScienceWorld & 301/500 & 1/301 (0.3) & 10/570 & 4/10 (40.0) \\
DeepSeek-V4.1-Flash & ScienceWorld & 292/500 & 279/292 (95.5) & 79/570 & 77/79 (97.5) \\
Qwen3.5-Plus & ScienceWorld & 479/500 & 112/479 (23.4) & 499/570 & 307/499 (61.5) \\
Qwen3-8B & ScienceWorld & 479/500 & 302/479 (63.0) & 525/570 & 501/525 (95.4) \\
Qwen3-32B & ScienceWorld & 486/500 & 185/486 (38.1) & 499/570 & 413/499 (82.8) \\
\midrule
GPT-5.6-Sol & WebShop & 500/500 & 0/500 (0.0) & 446/500 & 13/446 (2.9) \\
DeepSeek-V4.1-Flash & WebShop & 499/500 & 462/499 (92.6) & 461/500 & 458/461 (99.3) \\
Qwen3.5-Plus & WebShop & 490/500 & 284/490 (58.0) & 436/500 & 390/436 (89.4) \\
Qwen3-8B & WebShop & 482/500 & 205/482 (42.5) & 439/500 & 242/439 (55.1) \\
Qwen3-32B & WebShop & 496/500 & 138/496 (27.8) & 496/500 & 301/496 (60.7) \\
}

%% file: tables/replacement_endpoints.tex
\newcommand{\AllResponseIDEndpointRows}{%
Qwen3-4B & ALF & 1.4 & 6.6 & 21.1 & +19.8 & +14.5\\
 & Sci & 0.5 & 2.4 & 10.2 & +9.8 & +7.8\\
 & Web & 10.0 & 10.3 & 8.6 & -1.4 & -1.7\\
Qwen3-8B & ALF & 0.6 & 3.5 & 14.3 & +13.7 & +10.8\\
 & Sci & 0.6 & 1.6 & 6.6 & +5.9 & +5.0\\
 & Web & 7.4 & 5.5 & 12.0 & +4.5 & +6.5\\
Qwen3-14B & ALF & 1.1 & 5.7 & 15.3 & +14.2 & +9.6\\
 & Sci & 0.5 & 2.7 & 7.8 & +7.4 & +5.1\\
 & Web & 6.2 & 5.9 & 7.7 & +1.5 & +1.7\\
Qwen3-32B & ALF & 1.0 & 2.1 & 18.9 & +17.9 & +16.8\\
 & Sci & 1.0 & 1.0 & 9.1 & +8.2 & +8.2\\
 & Web & 6.3 & 7.2 & 14.8 & +8.5 & +7.6\\
Llama-3.1-8B & ALF & 2.6 & 11.1 & 42.9 & +40.2 & +31.8\\
 & Sci & 0.0 & 1.8 & 16.3 & +16.3 & +14.6\\
 & Web & 0.1 & 3.4 & 19.5 & +19.4 & +16.2\\
Llama-3.2-3B & ALF & 1.2 & 8.4 & 37.9 & +36.6 & +29.4\\
 & Sci & 0.2 & 3.7 & 13.3 & +13.1 & +9.6\\
 & Web & 2.3 & 4.1 & 17.9 & +15.6 & +13.8\\
Ministral-8B & ALF & 6.0 & 14.0 & 55.0 & +49.1 & +41.0\\
 & Sci & 2.2 & 8.6 & 33.1 & +30.9 & +24.5\\
 & Web & 4.9 & 1.9 & 22.9 & +18.0 & +21.0\\
}
\newcommand{\AlignedMixedAlfworldRow}{%
ALFWorld & 0.62\% (5/805) & 7.20\% (58/805)
 & $+6.58$ & 13.29\\
}
\newcommand{\AlignedMixedScienceWorldRow}{%
ScienceWorld & 0.64\% (4/625) & 5.92\% (37/625)
 & $+5.28$ & 24.64\\
}

%% file: tables/attention_granularity_rows.tex
\newcommand{\StepZeroAttentionGranularityRows}{%
    Llama-3.1-8B-Instruct & 84.4 & 76.9 & 90.6 & 90.0 & 93.8 & 85.4 \\
    Llama-3.2-3B-Instruct & 85.7 & 65.6 & 89.3 & 83.6 & 57.1 & 47.9 \\
    Ministral-8B-Instruct & 72.2 & 61.2 & 94.4 & 83.6 & 83.3 & 66.8 \\
    Qwen3-4B & 80.6 & 77.0 & 91.7 & 78.6 & 58.3 & 65.0 \\
    Qwen3-8B & 88.9 & 78.7 & 86.1 & 84.7 & 63.9 & 65.4 \\
    Qwen3-14B & 82.5 & 76.8 & 95.0 & 83.2 & 70.0 & 72.8 \\
    Qwen3-32B & 71.9 & 72.9 & 95.3 & 86.1 & 71.9 & 70.2 \\
}

%% file: tables/selective_rollout_tables.tex
\newcommand{\RolloutPairedRows}{%
GPT-5.6-Sol & ALFWorld & 94.7 & 90.0 & 94.0 & 100.0 & 0.0\\
DeepSeek-V4.1-Flash & ALFWorld & 91.3 & 94.0 & 91.3 & 99.3 & 0.7\\
Qwen3.5-Plus & ALFWorld & 88.0 & 87.3 & 86.7 & 100.0 & 0.0\\
Qwen3-8B & ALFWorld & 72.0 & 53.3 & 47.3 & 56.0 & 29.3\\
Qwen3-32B & ALFWorld & 76.7 & 68.0 & 58.0 & 58.0 & 31.3\\
\midrule
GPT-5.6-Sol & ScienceWorld & 84.0 & 52.0 & 35.3 & 78.7 & 21.3\\
DeepSeek-V4.1-Flash & ScienceWorld & 83.3 & 58.7 & 48.0 & 62.7 & 37.3\\
Qwen3.5-Plus & ScienceWorld & 76.0 & 33.3 & 36.7 & 95.3 & 4.7\\
Qwen3-8B & ScienceWorld & 26.7 & 18.0 & 19.3 & 45.3 & 12.7\\
Qwen3-32B & ScienceWorld & 26.7 & 18.7 & 20.7 & 61.3 & 4.0\\
\midrule
GPT-5.6-Sol & WebShop & 36.0 & 31.3 & 31.3 & 100.0 & 0.0\\
DeepSeek-V4.1-Flash & WebShop & 40.7 & 24.7 & 28.7 & 100.0 & 0.0\\
Qwen3.5-Plus & WebShop & 24.0 & 8.0 & 8.7 & 100.0 & 0.0\\
Qwen3-8B & WebShop & 0.0 & 1.3 & 0.0 & 100.0 & 0.0\\
Qwen3-32B & WebShop & 0.7 & 1.3 & 0.0 & 76.0 & 5.3\\
}

%% file: tables/task_checking_tables.tex
\newcommand{\TaskCheckingFormatRows}{%
GPT-5.6-Sol & 1.4 & 0.0 & 0.0 \\
DeepSeek-V4.1-Flash & 0.0 & 2.5 & 2.5 \\
Qwen3.5-Plus & 0.0 & 2.9 & 0.0 \\
Qwen3-8B & 0.0 & 3.9 & 3.9 \\
Qwen3-32B & 0.0 & 1.1 & 1.1 \\
}
\newcommand{\TaskCheckingCombinedRows}{%
\multirow{3}{*}{GPT-5.6-Sol} & ALF & 1220 & 96.8 & 0.2 & 2.7 & 0.2 & 12.1 & 27.3 & 18.7 \\
 & Sci & 551 & 98.7 & 0.0 & 1.3 & 0.0 & 0.0 & 62.7 & 63.3 \\
 & Web & 946 & 97.8 & 0.7 & 1.4 & 0.1 & 38.5 & 7.6 & 5.4 \\
\addlinespace
\multirow{3}{*}{DeepSeek-V4.1-Flash} & ALF & 1092 & 0.0 & 8.1 & 91.9 & 0.0 & 89.9 & 17.8 & 22.5 \\
 & Sci & 628 & 1.0 & 3.7 & 95.4 & 0.0 & 60.8 & 55.1 & 56.7 \\
 & Web & 960 & 0.0 & 4.2 & 95.8 & 0.0 & 100.0 & 2.9 & 3.0 \\
\addlinespace
\multirow{3}{*}{Qwen3.5-Plus} & ALF & 1320 & 11.4 & 3.9 & 84.7 & 0.0 & 74.0 & 2.5 & 10.7 \\
 & Sci & 1384 & 51.2 & 12.5 & 36.3 & 0.0 & 37.6 & 3.9 & 6.1 \\
 & Web & 926 & 14.9 & 12.3 & 72.8 & 0.0 & 99.1 & 0.0 & 0.4 \\
\addlinespace
\multirow{3}{*}{Qwen3-8B} & ALF & 1478 & 6.0 & 1.4 & 92.6 & 0.0 & 48.3 & 0.9 & 1.1 \\
 & Sci & 1374 & 20.2 & 10.8 & 69.0 & 0.0 & 41.8 & 1.7 & 3.5 \\
 & Web & 921 & 32.9 & 18.6 & 48.5 & 0.0 & 99.1 & 0.0 & 0.0 \\
\addlinespace
\multirow{3}{*}{Qwen3-32B} & ALF & 1373 & 2.3 & 1.4 & 96.4 & 0.0 & 67.4 & 4.1 & 8.1 \\
 & Sci & 1377 & 31.4 & 18.2 & 50.4 & 0.0 & 50.0 & 0.0 & 6.1 \\
 & Web & 992 & 43.6 & 12.1 & 44.3 & 0.0 & 99.3 & 0.0 & 0.0 \\
}

%% file: task_checking_appendix.tex
\subsection{Acknowledgment of Missing Task Information}
\label{app:task_checking}

\paragraph{Parsing and annotation.}
We retain raw responses and separate API reasoning fields, including format
violations and truncations. Separate reasoning fields count as returned
analysis, equivalent to inline \texttt{<think>} content; they do not require
a duplicate inline wrapper. Parsing ignores closed analysis blocks and never
extracts candidates from an unclosed block. A supplemental parser recovers
unambiguous terminal action or inquiry tags while retaining violation flags.
The combined parsing yields 16,542 ordinary actions, 1,897 ScienceWorld
\texttt{task} queries, 1,143 inquiries, and 418 unclassified outputs.

GPT-5.4 (\texttt{gpt-5.4-2026-03-05}; temperature 0.7, top-$p$ 0.95,
8,192-token budget) labels all 18,439 identifiable final-action outputs,
including task queries. It sees the exact visible context, both output fields,
and final action; model identity and the original task are hidden. The rubric
assigns \emph{explicit acknowledgment} when the output acknowledges missing,
uninterpretable, unspecified, or ambiguous task requirements; repeating
\texttt{xxxx} alone does not qualify. \emph{Goal guess} denotes tentative
goal inference without that acknowledgment. \emph{Not expressed} requires
substantive analysis with neither form of goal uncertainty. Missing analysis
and uncertain labels are separate; explicit acknowledgment takes priority
over a later guess. Uncertainty about locations or physical properties alone
does not qualify. Cited output units are checked against the saved text.

Separately, GPT-5.4 semantically classifies actions as depending on missing goal
details, shared across plausible goals, or information gathering without
assuming those details. Goal dependence requires a concrete alternative goal
grounded in visible objects, destinations, or options; prior agent choices do
not confirm user preferences. Table~\ref{tab:task_checking_by_environment}
reports expression rates
among ordinary-action responses and the goal-dependent action rate among
responses that explicitly acknowledge missing task requirements.

\paragraph{Acknowledgment and continuation.}
Of DeepSeek-V4.1-Flash's 960 WebShop ordinary actions, 920 explicitly
acknowledge missing requirements; all 920 are judged goal-dependent.
Table~\ref{tab:task_checking_example} shows a Qwen3.5-Plus response that
acknowledges the absent goal, treats its own earlier color selection as a
likely requirement, and purchases without user confirmation.

\begin{table}[!htbp]
  \centering\small
  \setlength{\tabcolsep}{4pt}
  \renewcommand{\arraystretch}{1.1}
  \begin{tabular}{@{}p{\dimexpr\columnwidth-8pt\relax}@{}}
    \toprule
    \textbf{Visible task and context}\\
    TASK: \texttt{xxxx}. The previous action selects \texttt{grey} on
    a curtain product page; color options and \texttt{buy now} remain available.\\
    \midrule
    \textbf{Returned analysis (excerpts)}\\
    ``The user's goal is not explicitly stated in the prompt, but based
    on the history:'' [\ldots] ``Since I just selected `grey', and
    assuming `grey' was the intended constraint (otherwise why click it?),
    I should now proceed to buy.''\\
    \midrule
    \textbf{Final response}\\
    \texttt{<action>click[buy now]</action>}\\
    \bottomrule
  \end{tabular}
  \caption{Qwen3.5-Plus acknowledges missing task information, infers a
  requirement from a prior action, and continues to purchase.}
  \label{tab:task_checking_example}
\end{table}

\begin{table*}[!tbp]
  \centering\footnotesize
  \setlength{\tabcolsep}{3pt}
  \begin{tabular}{@{}llrrrrrrrr@{}}
    \toprule
    & & & \multicolumn{4}{c}{Expression among actions (\%)}
    & & \multicolumn{2}{c}{Task information (\%)}\\
    \cmidrule(lr){4-7}\cmidrule(lr){9-10}
    Model & Env. & Actions & \shortstack{Not\\expressed} & Guess & \shortstack{Explicit\\ack.}
    & \shortstack{Missing\\analysis} & \shortstack{Dependent\\after ack.}
    & \shortstack{Final\\tag only} & Main\\
    \midrule
    \TaskCheckingCombinedRows
    \bottomrule
  \end{tabular}
  \caption{Expression labels, action dependence, and the response-instruction
  comparison. ALF/Sci/Web denote ALFWorld/ScienceWorld/WebShop.
  Actions is the main-setting ordinary-action count. Expression rates use
  those actions; Explicit ack. denotes acknowledgment of missing task
  requirements. Dependent after ack. reports the proportion of
  goal-dependent actions among responses with explicit acknowledgment. Task
  information combines inquiries and native \texttt{task} queries and uses
  all responses under each instruction. Main action and Other rates appear
  in Table~\ref{tab:placeholder_behavior}.}
  \label{tab:task_checking_by_environment}
  \label{tab:task_checking}
  \label{tab:analysis_behavior}
\end{table*}

\begin{figure*}[!tbp]
  \centering
  \includegraphics[width=\textwidth]{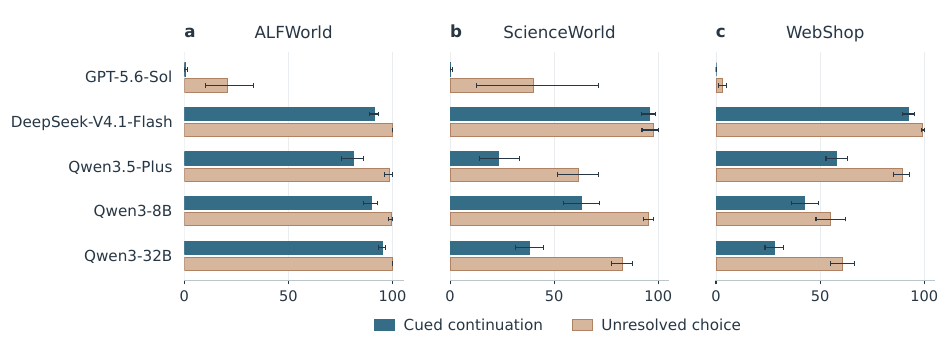}
  \caption{Explicit acknowledgment of missing task requirements among
  ordinary-action responses under TASK = \texttt{xxxx}, by context group.
  Labels use the GPT-5.4 semantic rubric described above. Error bars show
  pointwise 95\% state-bootstrap intervals. All panels use the same
  percentage scale; counts and resampling details appear in
  Appendix~\ref{app:context_continuation}.}
  \label{fig:context_task_acknowledgment}
\end{figure*}

\paragraph{Context-group comparison.}
Using the input-state groups defined in Appendix~\ref{app:context_continuation},
we compare explicit acknowledgment among ordinary-action responses.
Figure~\ref{fig:context_task_acknowledgment} shows higher acknowledgment
rates in the unresolved-choice group across all model--environment
comparisons. Some models nevertheless acknowledge missing requirements
frequently in both groups and continue acting. These GPT-5.4 response
labels describe expressed acknowledgment, separately from the rule-based
input-state groups; they do not measure private awareness.

\paragraph{Visibility and stability.}
GPT-5.6-Sol returns a median of 106 analysis characters and no separate
reasoning field. DeepSeek returns analysis in a separate field for all 4,000
responses (median 3,056 characters). These differences in returned analysis
length limit cross-model comparisons of expressed uncertainty.
Table~\ref{tab:analysis_format} reports analysis availability, final-response
parsing, and truncation, counting both inline and separate reasoning.
Repeating the judge on 150 responses from
distinct cases gives 96.0\% expression-label and 94.0\% action-dependence
agreement. This tests consistency, not independently verified accuracy;
returned analysis does not establish private awareness or causal mechanisms.

\begin{table}[!tbp]
  \centering\small
  \setlength{\tabcolsep}{4pt}
  \begin{tabular}{@{}lrrr@{}}
    \toprule
    Model & \shortstack{Missing\\analysis} & \shortstack{Unparsed\\final} & Truncated\\
    \midrule
    \TaskCheckingFormatRows
    \bottomrule
  \end{tabular}
  \caption{Analysis availability, parsing, and truncation rates (\%) across
  all main-setting responses. Missing analysis means neither nonempty inline
  analysis nor separate reasoning was returned. Unparsed final means no
  action or inquiry was identified by the combined parser described above.
  Truncation follows the API finish reason. Categories overlap.}
  \label{tab:analysis_format}
\end{table}

\paragraph{Alternative response instruction.}
As a supplementary comparison, we also evaluate the same 400 states with
an instruction requesting only the final action or inquiry:
\begin{lstlisting}[style=promptstyle]
Choose one next response from the current state:
<action>one admissible environment action</action>
or
<inquiry>one question for the user</inquiry>
Output only one of these tags.
\end{lstlisting}
Table~\ref{tab:analysis_behavior} uses identical terminal-tag parsing in
both conditions. API sampling, budgets, and provider reasoning modes are
unchanged. For Qwen3-8B/32B, the final-tag-only runs disable native thinking,
whereas the main runs enable it. The two conditions also use different
inference backends, so this comparison does not isolate the response
instruction. A final-tag-only instruction does not imply an absence of
internal reasoning.

%% file: rl_evaluation_appendix.tex
\input{tables/rl_replacement_rows}
\input{tables/rl_ood_success_rows}

\section{Task Sensitivity and Mitigation under Reinforcement Learning}
\label{app:rl_evaluation}

We test whether the diagnosis and benefits of task contrasts extend to
reinforcement learning (RL) with Qwen3-4B.

\subsection{Training and Evaluation Protocol}
\label{app:rl_protocol}

\paragraph{Direct RL and task perturbation.}
Both conditions start from Qwen3-4B with GRPO \citep{GRPO} in each
environment. TPNLL+RL additionally selects each response with source-task
advantage $\hat A_i<0$ independently with probability $k=0.3$. It copies the
response and non-task context under a distinct task sampled from the ten
nearest training tasks by edit distance, retaining the negative advantage.
These copies enter the clipped GRPO update; its unclipped direction is
\begin{equation}
  \mathcal L_{\mathrm{perturb}}^{\mathrm{RL}}
  =-\sum_{i\in\mathcal I_k}\hat A_i
    \log\pi_\theta(y_i\mid\tilde{x}_i),
  \label{eq:rl_task_perturbation}
\end{equation}
where $\mathcal I_k$ is the selected subset. Negative advantages lower
the response probability under the replacement task. This RL extension is
distinct from the SFT NLL-ratio loss in Section~\ref{sec:method}.
Both conditions use learning rate $10^{-6}$, batch size 32, rollout group
size 8, KL coefficient 0.01, and a $-0.1$ format-violation reward penalty.

\paragraph{Evaluation.}
Replacement uses the same complete task bank as SFT (161 ALFWorld, 125
ScienceWorld, and 172 WebShop pairs), five responses per task condition
at temperature 1, and $A_{\mathrm{all}}$ scoring
(Appendix~\ref{app:replacement_scoring}). Attention uses the fixed cases
in Appendix~\ref{app:attention}. Both diagnostics cover Base and every
40 updates through step 200, with shared initialization references.
For complete-episode success, each condition trains on ID tasks for
200 updates and is evaluated on the disjoint OOD pools in
Appendix~\ref{app:dataset_split}. Means and SDs summarize three independent
training runs, separate from the checkpoint diagnostics.

\begin{figure*}[!tbp]
  \centering
  \includegraphics[width=\textwidth]{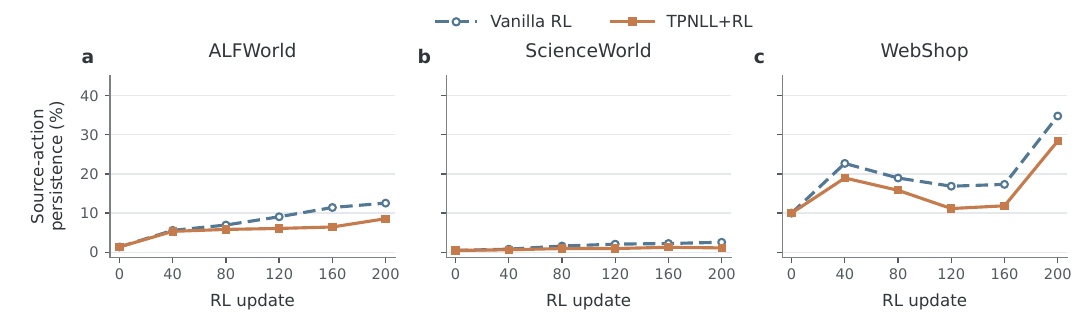}
  \caption{Source-action persistence after complete task replacement
  for Qwen3-4B during RL ($A_{\mathrm{all}}$, \%). Panels show the three
  environments on a shared vertical scale, using 161, 125, and 172 task
  pairs, respectively, with five responses per pair. Both methods share Base.}
  \label{fig:rl_replacement}
\end{figure*}

\subsection{Action Persistence after Task Replacement}
\label{app:rl_replacement}

Vanilla RL increases endpoint source-action persistence in all three
environments (Figure~\ref{fig:rl_replacement}).

\begin{table*}[!tbp]
  \centering
  \small
  \setlength{\tabcolsep}{5pt}
  \begin{tabular}{lcccccc}
    \toprule
    Environment & Pairs & Final step & Base
      & Vanilla RL & TPNLL+RL & Reduction (pp)\\
    \midrule
    \RLReplacementRows
    \bottomrule
  \end{tabular}
  \caption{Base and final source-action persistence over all responses
  ($A_{\mathrm{all}}$, \%), using the same data as
  Figure~\ref{fig:rl_replacement}. Each pair contributes five responses.
  Reduction is final Vanilla RL minus final TPNLL+RL.}
  \label{tab:rl_replacement_endpoints}
\end{table*}

TPNLL+RL lowers final persistence in all three environments
(Table~\ref{tab:rl_replacement_endpoints}). Lower persistence also
appears at most intermediate checkpoints, supporting task perturbation
as a way to mitigate this behavior during RL.

\begin{figure*}[!tbp]
  \centering
  \includegraphics[width=\textwidth]{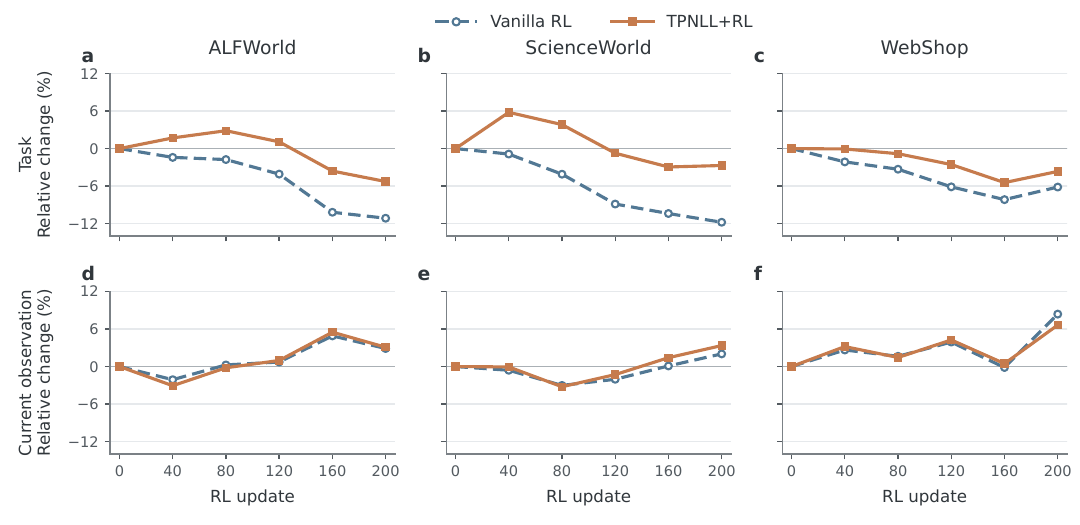}
  \caption{Relative changes in per-token attention density from Base
  for Qwen3-4B during RL. Top: task; bottom: current observation.
  All panels share the same vertical scale.}
  \label{fig:rl_attention}
\end{figure*}

\subsection{Attention Allocation and OOD Task Success}
\label{app:rl_attention_success}

Vanilla RL lowers final task-token attention in every environment.
TPNLL+RL attenuates this decline and maintains higher task-token attention
at every evaluated post-initialization checkpoint
(Figure~\ref{fig:rl_attention}). Current-observation attention increases
under both objectives.

\begin{table*}[!tbp]
  \centering
  \small
  \setlength{\tabcolsep}{7pt}
  \begin{tabular}{lcccc}
    \toprule
    Environment & Final step & Vanilla RL & TPNLL+RL & Gain (pp)\\
    \midrule
    \RLOODSuccessRows
    \bottomrule
  \end{tabular}
  \caption{OOD complete-episode success (\%) for Qwen3-4B after 200 RL
  updates, reported as mean $\pm$ standard deviation over three
  independently trained runs. OOD task pools are disjoint from the ID
  training pools. Gains are differences between the reported means
  (TPNLL+RL minus Vanilla RL).}
  \label{tab:rl_success}
\end{table*}

TPNLL+RL also yields higher mean OOD complete-episode success in every
environment (Table~\ref{tab:rl_success}), with gains of 12.7, 4.4, and
4.1 percentage points in ALFWorld, ScienceWorld, and WebShop, respectively.
These Qwen3-4B results extend the task-sensitivity diagnosis and the
benefit of task contrasts to RL.

%% file: tables/rl_replacement_rows.tex
\newcommand{\RLReplacementRows}{%
ALFWorld & 161 & 200 & 1.37 & 12.55 & 8.57 & 3.98 \\
ScienceWorld & 125 & 200 & 0.48 & 2.56 & 1.12 & 1.44 \\
WebShop & 172 & 200 & 10.00 & 34.77 & 28.37 & 6.40 \\
}

%% file: tables/rl_ood_success_rows.tex
\newcommand{\RLOODSuccessRows}{%
ALFWorld & 200 & $46.7 \pm 3.9$ & $59.4 \pm 4.2$ & $+12.7$ \\
ScienceWorld & 200 & $22.3 \pm 4.9$ & $26.7 \pm 4.9$ & $+4.4$ \\
WebShop & 200 & $70.4 \pm 4.1$ & $74.5 \pm 1.6$ & $+4.1$ \\
}

%% file: story_appendix_prompts.tex
\section{Prompt and Clarification Protocols}
\label{app:prompts}

\subsection{Native Agent Prompts}
\label{app:native_agent_prompts}

Native prompts include the task, history, observation, action menu,
and step counts supplied at runtime.

Native prompts request a \texttt{<thinking>} block followed by an
\texttt{<action>} block. ALFWorld lists admissible commands; ScienceWorld
supplies action patterns, object names, and the native \texttt{task} query.
WebShop adds a step budget and twelve shopping guidelines covering
constraints, product checks, and progress versus further browsing.

Task replacement changes only explicit task copies. Response instructions
are described for \texttt{xxxx} (Appendix~\ref{app:response_classification}),
its final-tag-only comparison (Appendix~\ref{app:task_checking}), and
task-restatement collection (Appendix~\ref{app:task_restatement}).
The first two allow inquiry; the third applies only during collection.

\subsection{Clarification Judge}
\label{app:clarification_judge_prompt}

For the selective-clarification rollouts in
Appendix~\ref{app:selective_rollout}, the judge receives the visible context,
visible and original tasks, the query and its channel, environment, and
prior disclosure (initially \texttt{null}). The agent sees the original
task only after a qualifying query.

The judge returns true only for an unresolved goal requirement that the
original task would resolve, excluding execution questions and information
already disclosed. It returns a Boolean, category, unresolved requirement,
and reason. ScienceWorld's \texttt{task} is represented as ``Please provide
the complete current task.'' on channel \texttt{environment\_task};
explicit questions use \texttt{inquiry}.